\PassOptionsToPackage{sort&compress}{natbib}
\documentclass{article}

\usepackage[preprint]{corl_2026} 

\title{\ourframework: Policy-Agnostic Real-World RL for Tactile Residual Adaptation of Visual Policies}

\definecolor{nicegreen}{rgb}{0.1, 0.6, 0.2}

\author{
  Kelin Yu$^{1*}$ \quad Haode Zhang$^{1*}$ \quad Harish Ravichandar$^{2}$ \quad Yunhai Han$^{2}$ \quad Ruohan Gao$^{1}$\\[-1pt]
  $^{1}$University of Maryland, College Park \quad $^{2}$Georgia Institute of Technology\\[-1pt]
  $^{*}$Equal contribution\\[-1pt]
  {\small Project page: \url{https://colinyu1.github.io/omnitactune-site/}}
}
\usepackage{amsmath,amssymb,amsthm}
\usepackage{amsfonts} 

\usepackage{graphicx}
\usepackage{booktabs}
\usepackage{array}
\usepackage{multirow}
\usepackage{booktabs}
\usepackage[table]{xcolor}
\usepackage{caption}
\usepackage{xspace}
\usepackage{url}

\def\ourframework{\textsc{OmniTacTune}\xspace}

\begin{document}
\maketitle

\vspace{-25pt}
\begin{abstract}
    Visual policies learned from human videos, teleoperation, and robot demonstrations offer scalable motion priors, but often fail in contact-rich manipulation, where success significantly depends on local force and contact geometry. Tactile sensing provides these complementary signals, yet tactile data remain 
    costly to collect and hard to generalize across sensors, robots, and tasks. We introduce \ourframework, a policy-agnostic real-world RL pipeline that adapts tactile feedback to pretrained visual policies through residual correction. \ourframework uses a two-stage design: it first warm-starts tactile-aware learning from autonomous base-policy rollouts, then learns a lightweight tactile residual policy through online interaction. 
    Extensive experiments show that \ourframework generalizes across diverse contact-rich tasks, visual base policies, 
    and tactile representations. Across four real-world contact-rich tasks, it improves visual base policies from 5--40\% success to 85--100\% within 40--80 minutes, demonstrating an efficient path for adapting tactile feedback to scalable visual robot policies.
\end{abstract}
\vspace{-0.12in}

\begin{figure*}[!h]
    \centering
    \vspace{-6pt}
    \includegraphics[width=\textwidth]{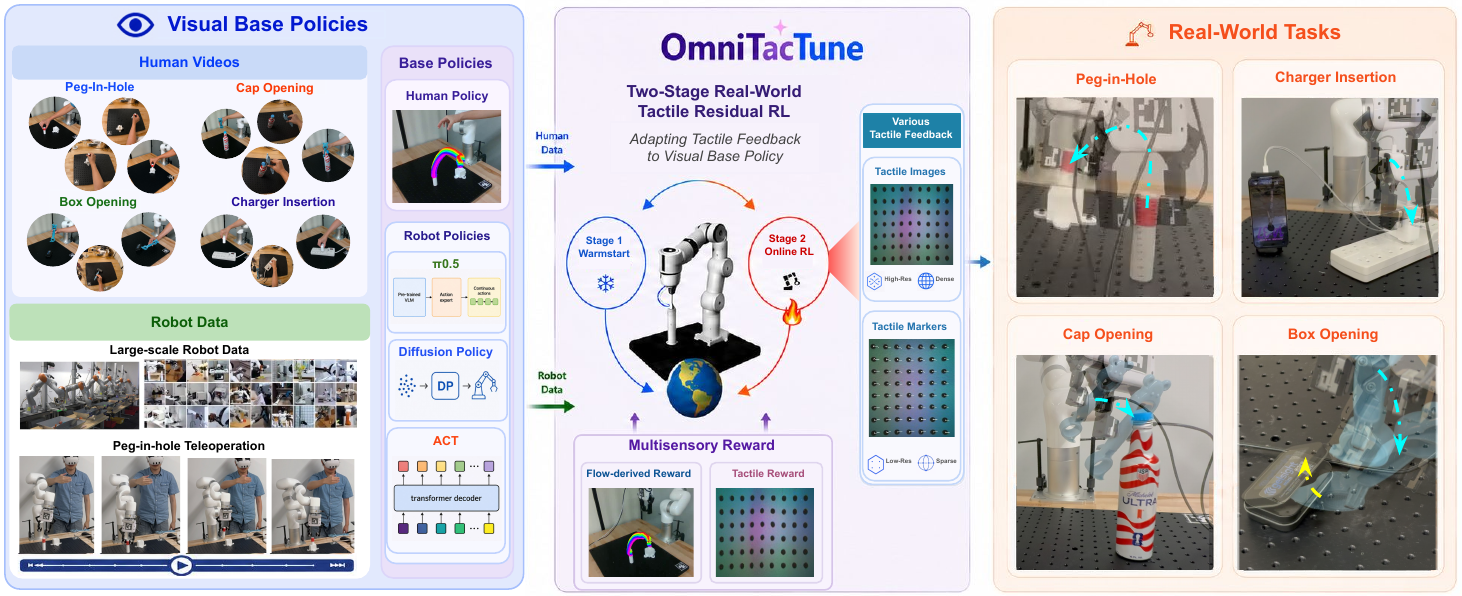}
    \caption{\textbf{\ourframework{}} adapts tactile feedback to diverse visual base policies trained from human videos or robot data~(\textbf{left}) through a two-stage real-world tactile residual RL pipeline~(\textbf{middle}), 
    enabling efficient tactile adaptation across challenging contact-rich manipulation tasks (\textbf{right}).
    }
    \label{fig:teaser}
    \vspace{-17pt}
\end{figure*}
\section{Introduction}
\vspace{-8pt}

Visual robot policy learning has made tremendous progress, driven in large part by scaling. The field is moving toward increasingly scalable data sources, from teleoperated robot demonstrations~\cite{openxembodiment, droid} to more recently large-scale human videos~\cite{egomimic, egoscale, egoverse}, offering a promising path toward general visual policies across diverse tasks, environments, and embodiments.
However, scaling visual data alone does not resolve the challenges of contact-rich manipulation. Cameras provide rich spatial observations for planning, but they cannot directly measure contact or forces during physical interaction, such as insertion and tool use. As a result, many failures occur precisely at contact, where unobserved physical interactions determine whether a task succeeds or fails.


Touch, on the other hand, provides direct access to local interaction signals critical for contact-rich manipulation, including normal force, shear force, and contact geometry~\cite{gelsight, mimictouch}. Prior work has shown that tactile sensing improves tasks requiring precise contact reasoning~\cite{3dvitac, mimictouch, seehearfeel, rdp}, 
but these methods often rely on task-specific tactile data collected with a particular sensor or setup~\cite{tarf, touchinthewild, tvl, objbench}, limiting cross-platform transfer. 
Despite recent efforts to scale tactile data~\cite{touchinthewild, vitamin,objbench}, tactile datasets remain orders of magnitude smaller than visual datasets~\cite{openxembodiment,egoscale, egodex}. 
This creates a fundamental scale gap, as tactile observations are rarely paired with the large-scale visual data from which general robot behaviors are increasingly learned.

Our key idea is to leverage the best of both worlds, rather than scale tactile data to the level of vision. Vision and touch play different roles in manipulation, much like in human learning. By watching others, humans can infer the global structure of a task: what objects are involved, how they move, and what action sequence is likely to succeed. Yet reliable execution often comes only through hands-on practice, where touch reveals whether parts are aligned, resistance is too high, and small adjustments are needed. This distinction suggests a scalable paradigm for contact-rich manipulation: learn general visual policies from abundant visual data, then use tactile feedback to refine and correct these policies during contact. In this view, vision provides broad, scalable supervision for task-level behavior, while touch provides local, residual feedback for the last mile of physical interaction.

Learning such tactile residual corrections is challenging because the necessary supervision is revealed only through physical interaction. 
Prior real-to-sim-to-real approaches~\cite{vtrefine, xsim, lodestar} are poorly suited to this setting, as they must bridge visual, dynamics, and tactile gaps simultaneously, while tactile signals and dynamic interactions are especially difficult to simulate. 
Imitation-based correction methods, such as DAgger-style approaches~\cite{dagger}, can correct policy failures through expert interventions, but they still depend on repeated human demonstrations rather than autonomous improvement. We therefore turn to \emph{real-world reinforcement learning (RL)}~\cite{serl, mimictouch, pld, effrl, rlt} as a natural mechanism for learning tactile residuals. Continuing the human-learning analogy, we propose to use visual priors to guide the overall behavior and tactile feedback to refine execution through real-world practice, enabling the robot to improve its contact strategy from experience.

\begin{figure*}[!t]
    \centering
    \includegraphics[width=\textwidth]{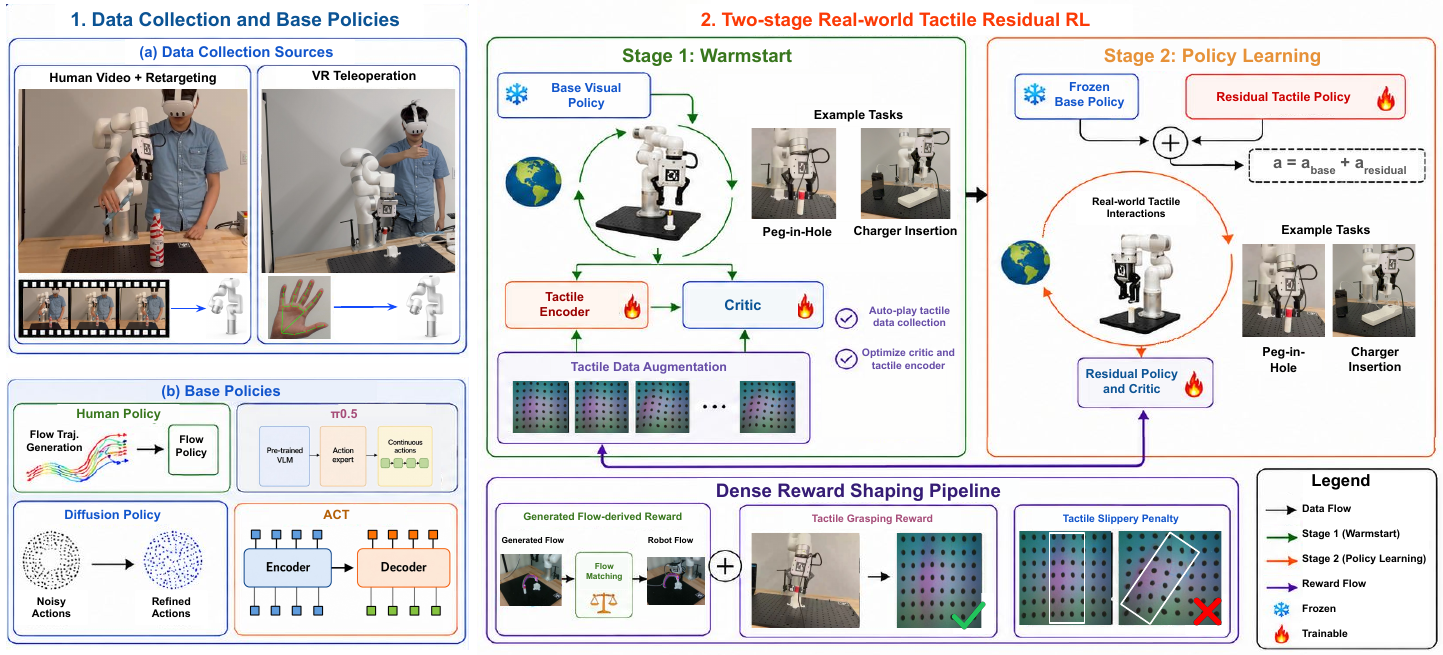}
    \vspace{-15pt}
    \caption{\textbf{System overview of \ourframework{}.}
    \ourframework{} first collects visual demonstrations via human video retargeting and VR teleoperation, and uses them to train diverse visual base policies~(\textbf{left}). It then adapts tactile feedback through a two-stage real-world residual RL pipeline: a warm-start stage collects tactile rollouts to optimize the tactile encoder and critic, followed by an online policy learning stage that learns residual tactile corrections~(\textbf{top right}). A dense reward shaping pipeline combines visuo-tactile feedback to guide efficient real-world learning~(\textbf{bottom}).}
    \label{fig:pipeline}
    \vspace{-15pt}
\end{figure*}

Towards this goal, we introduce \textbf{\ourframework} (Fig.~\ref{fig:pipeline}), a two-stage real-world RL pipeline for tactile residual adaptation across visual base policies. Firstly, \ourframework leverages a pretrained visual base policy, learned from scalable visual data such as human videos or robot demonstrations, as task-level motion priors. In the \emph{first stage}, autonomous rollouts from base policies enable warm-start training, which initializes the replay buffer, bootstraps the critic, and adapts the tactile representation to the target task. In the \emph{second stage}, \ourframework performs online residual RL to learn a tactile policy that predicts contact-aware corrections on top of the frozen visual policy. To improve sample efficiency and stability, we further introduce object-centric visuo-tactile reward shaping, which provides dense feedback and guides exploration during real-world practice. Together, this framework enables reliable residual adaptation from real-world interaction.


Our main contributions are threefold. \emph{First}, we formulate tactile adaptation as residual correction on top of existing visual policies, enabling tactile-aware practice without training from scratch. \emph{Second}, we introduce \ourframework, a plug-and-play two-stage real-world RL pipeline for online tactile residual adaptation, together with a multi-sensory reward that combines generated object-centric guidance with tactile contact signals for sample-efficient real-world refinement.
\emph{Third}, we demonstrate that \ourframework generalizes across contact-rich tasks, visual base policies, and tactile representations. On four challenging real-world contact-rich tasks, \ourframework consistently improves success rates from 5--40\% to 85--100\% within 40--80 minutes of online practice.




\vspace{-4pt}
\section{Related Work}

\paragraph{Visual Policy Learning from Robot and Human Data.}
\vspace{-0.2cm}
Recent robot learning methods increasingly leverage scalable visual data, including robot demonstrations, teleoperation data, and human videos, to learn general manipulation policies~\cite{openxembodiment, droid, act, dp, egomimic, im2flow2act, egoscale}. 
For robot data, imitation-learning backbones such as ACT~\cite{act}, Diffusion Policy~\cite{dp}, and recent vision-language-action models~\cite{pi05, pi06, openvla} have shown strong performance when sufficient robot demonstrations are available. 
For human videos, prior work reduces the embodiment gap through human-robot co-training~\cite{egomimic, emma, immimic}, object-centric representations~\cite{im2flow2act, genflowrl, cimer, humanego, rekep, afford}, or interaction-based policy refinement~\cite{humanwild, mimictouch}. 
However, these visual policies still lack direct contact feedback, and we use them as frozen base policies while learning tactile residual corrections for contact-rich execution.
\vspace{-12pt}
\paragraph{Tactile and Multi-Sensory Robot Manipulation.}
Tactile sensing provides local contact information that is difficult to infer from RGB observations, such as contact geometry and slip~\cite{gelsight, 3dvitac, anyskin, digit}, and has benefited a wide range of contact manipulation tasks, 
including grasping, insertion, in-hand manipulation, and dexterous manipulation~\cite{seehearfeel, actcontact, grasp, letacmpc, touchdex, tdex, rdp, factr}. 
Prior work incorporates tactile feedback through
visuo-tactile sensory fusion~\cite{seehearfeel}, slow-fast multimodal architectures~\cite{rdp}, tactile world model~\cite{letacmpc}, and reinforcement learning~\cite{touchdex, powerofsense}. However, most existing visuo-tactile policy learning methods require paired, task-specific visuo-tactile demos to learn policies. In contrast, \ourframework adapts tactile feedback online as residual corrections on top of existing visual policies, enabling contact-aware practice without training a visuo-tactile policy from scratch.

\vspace{-12pt}
\paragraph{Policy Adaptation for Contact-Rich Manipulation.}
Policy adaptation methods improve pretrained policies through additional interaction. DAgger-style methods~\cite{dagger, hgdagger, crdagger} reduce compounding errors with expert corrections during execution, and CR-DAgger~\cite{crdagger} further adapts force-reactive behaviors, but these methods still rely on repeated human intervention. Real-to-sim-to-real methods~\cite{lodestar, xsim, vividex, vtrefine} aim to reduce real-world data collection by adapting policies through simulation, but contact-rich manipulation is especially sensitive to simulation gaps in dynamics, contact, and tactile sensing. Real-world RL avoids these gaps by learning directly on hardware, with prior work improving sample efficiency through offline expert demos, pretrained policies, or warm-start rollouts~\cite{serl, offlinerl, qlofflinerl, cal-ql, rlpd, effrl, rlt, pld, mimictouch, im2refine, fish}. Our method also uses real-world RL, but addresses a different adaptation setting: \ourframework adapts a visual base policy with newly introduced tactile feedback. 
 
\vspace{-8pt}
\section{Methodology}
\vspace{-6pt}\label{sec:method}
We aim to build a policy-agnostic real-world RL pipeline for residual tactile adaptation across different visual base policies, tasks, and tactile representations. This setting introduces several bottlenecks that are not addressed by prior real-world RL pipelines~\cite{pld, serl}. Offline demonstrations often provide only visual feedback, and therefore cannot directly initialize a tactile replay buffer or tactile-aware critic. Warm-start rollouts can collect on-policy data, but usually do not explicitly bootstrap the critic or optimize the tactile encoder. This is particularly limiting because  pretrained tactile encoders can exhibit substantial representation shift across tasks and contact conditions. 

To address these challenges, we introduce \ourframework (see Fig.~\ref{fig:pipeline}).  We first describe how we collect human videos and robot data to train different visual base policies (Sec.~\ref{sec:data} and Sec.~\ref{sec:base}). We then present our two-stage real-world RL pipeline (Sec.~\ref{sec:residual}): during warm-start, the robot autonomously rolls out the frozen base policy to initialize the replay buffer, bootstrap the flow-tactile critic, and fine-tune the tactile encoder; during online training, residual RL learns tactile-informed corrections for the frozen visual policy. Finally, we introduce an object-centric multi-sensory reward that enables more stable and efficient RL training (Sec.~\ref{sec:reward}). Details are shown in App.~\ref{app:implementation}.
 
 

\vspace{-8pt}
\subsection{Data Collection}
\vspace{-6pt}\label{sec:data}
\paragraph{Human Hand Demonstrations.}Similar to HUDOR~\cite{hudor}, we collect human demonstrations with a third-view RGB camera and a Meta Quest headset. 
The camera records in-scene videos of humans expert demonstrations, while the Quest tracks the 6-DoF hand pose under its camera frame${}^{q}\mathbf{T}_{h}(t)$. 
We use tabletop Aruco markers and the third-view camera to retarget it to the robot base frame via the third-view camera, obtaining robot-aligned hand trajectories as ${}^{r}\mathbf{T}_{h}(t) = {}^{r}\mathbf{T}_{w} {}^{w}\mathbf{T}_{q} {}^{q}\mathbf{T}_{h}(t)$, where ${}^{w}\mathbf{T}_{q}$ maps the Quest tracking frame to the world frame, ${}^{r}\mathbf{T}_{w}$ maps the world frame to the robot base frame, and ${}^{r}\mathbf{T}_{h}(t)$ is the resulting robot-frame hand pose. The resulting synchronized videos and trajectories are used to extract object-centric motion priors and train the base policies.
\vspace{-6pt}

\paragraph{Teleoperation Data Collection.}In addition to hand demonstrations, we collect robot demonstrations using a Meta Quest headset via OpenTeach~\cite{openteach}. The OpenTeach maps the tracked human hand motion from quest to robot end-effector poses. We record synchronized RGB observations, robot states, and actions, which are used to train robot-data-based visual base policies.
\vspace{-8pt}

\subsection{Learning Visual Base Policies}
\label{sec:base}
\vspace{-6pt}
In this section, we introduce four different base policies, where all of them can be used for residual tactile adaptations with our real-world RL pipeline.
\vspace{-6pt}
\paragraph{Flow-based Policy.}Our flow-based policy builds on top of Im2Flow2Act~\cite{im2flow2act}, GenFlowRL~\cite{genflowrl}, and Dex4D~\cite{dex4d}, where we want to enable cross-embodiment visual policy learning from human videos, aiming to transfer task-level motion priors from videos to robot manipulation. Given a demonstration video, we extract and track a sparse set of object keypoints using DINOv2~\cite{dinov2} and SAM~\cite{sam}, and then using CoTracker 3~\cite{cotracker3} to track their motions. Following Im2Flow2Act~\cite{im2flow2act}, we fine-tune a pretrained flow generator using our collected task-specific human or robot demonstrations to predict a complete task-level object flow from the initial observation. The generated flow provides two complementary signals for our system: an object-conditioned motion guidance for both base policy and residual policy, and a dense object-centric reward signal during residual RL.

To train the lightwight, embodiment-agnostic base policy, we convert the generated and observed object flows into compact policy inputs.  At time $t$, the flow-guided base policy is formulated as $\mathbf{a}_{t:t+K}^{\mathrm{b}}=\pi_{\theta}^{\mathrm{flow}}(\mathbf{o}_t^{\mathrm{flow}}, \mathbf{q}_t)$, where $\mathbf{q}_t$ is the robot proprioceptive state and $\mathbf{o}_t$ is the flow representation, $\mathbf{o}_t^{\mathrm{flow}}=[\mathbf{c}_0^{3D},\mathbf{c}_t,\mathbf{T}_{t_0\rightarrow t}^{\mathrm{rel}},\hat{\mathbf{c}}_{\ell},\hat{\mathbf{T}}_{t_0\rightarrow \ell}^{\mathrm{rel}},\mathbf{T}_{t\rightarrow \ell}^{\mathrm{rel}}]$. Here, $\mathbf{c}_0^{3D}$ is the initial 3D centroid of keypoints for grasping, while $\mathbf{c}_t$ and $\hat{\mathbf{c}}_{\ell}$ are the current and generated lookahead centroids of keypoints. The relative transformations $\mathbf{T}$ represent the observed object motion, generated subgoal motion, and remaining motion to the subgoal. The predicted actions $\mathbf{a}_{t:t+K}^{\mathrm{b}}$ are a sequence of $6D$ delta poses with binary grasping signal. We keep the generated goal fixed until the observed centroid is sufficiently close to it, where it enables the robot to reach the next subgoal with a closed-loop manner.
\vspace{-6pt}
\paragraph{Other Visual Base Policies.}We also instantiate \ourframework with Action Chunking Transformer (ACT)~\cite{act}, Diffusion Policy (DP)~\cite{dp}, and $\pi_{0.5}$~\cite{pi05}. All three policies are trained or fine-tuned on our teleoperation data with synchronized RGB observations, robot states, and actions.

\vspace{-5pt}
\subsection{Real-World Tactile-Informed Residual RL Adaptation}
\label{sec:residual}
\vspace{-5pt}
We introduce our two-stage real-world RL pipeline for adapting tactile feedback into different kinds of frozen visual base policies with high stability and efficiency. 

\vspace{-5pt}
\paragraph{Stage I: warm-start.}
A key challenge in our setting is enabling tactile residual learning without offline visual-tactile demonstrations. A naive approach~\cite{serl,touchbegin} is to roll out the base policy and start SAC on the collected transitions. However, this leads to unstable exploration: a randomly initialized critic produces unreliable Q-value estimates, while a pretrained tactile encoder suffers from representation shift. Our warm-start stage addresses both issues jointly. Rather than only populating the replay buffer with autonomous rollouts~\cite{effrl,pld}, we use these transitions to bootstrap a flow-tactile critic $Q_{\eta}(\mathbf{q}_t,\mathbf{z}^f_t,\mathbf{z}^{\tau}_t,\mathbf{a}_t)$, where $\mathbf{q}_t$ is proprioception, $\mathbf{z}^f_t$ is flow features, $\mathbf{z}^{\tau}_t$ is tactile features, and $\mathbf{a}_t$ is final action. At the same time, we adapt a pretrained tactile encoder, such as AnyTouch2~\cite{anytouch2}, to the specific task. The tactile encoder is optimized with both critic loss and reconstruction regularization~\cite{powerofsense}, $\mathcal{L}_{E}=\mathcal{L}_{Q}+\lambda_{\text{rec}}\mathcal{L}_{\text{rec}}$. We only use contact transitions for optimizing tactile encoder, where contact is detected by thresholding marker displacement. By optimizing the tactile encoder and the critic, the online tactile residual learning can be more stable and efficient.

To improve data efficiency, inspired by DrqV2~\cite{drqv2}, we conduct trajectory-level tactile augmentation after each rollout. For each collected trajectory, we synthesize two new trajectories with temporally consistent tactile images conditioned on contact force while keeping the robot states, actions, and reward labels unchanged. These augmented transitions provide additional contact-rich samples for initializing the tactile representation and flow-tactile critic before online residual learning. We utilize ControlTac~\cite{controltac}-style SOTA augmentation method, with more details shown in Sec.~\ref{app:controltac}.

\vspace{-8pt}
\paragraph{Stage II: online tactile residual learning.}
The key question here is how a single residual actor can correct architecturally diverse policies such as Flow Policy~\cite{im2flow2act} and ACT~\cite{act}. We address this through interface-level agnosticism: instead of relying on internal representations, the residual actor uses two policy-independent inputs: the generated keypoints goals, which provide a shared task-level guidance, and the base-policy action chunk, which exposes its short-horizon motion intent. This allows the same residual actor to be attached to different visual policies for closed-loop correction. Specifically, the final action is $\mathbf{a}_t=\mathbf{a}_t^b+\mathbf{s}_t\mathbf{a}_t^r$, where $\mathbf{a}_t^b$ is the base action, $a_t^r$ is the residual correction, and $s_t$ is the scheduler. The residual policy takes as input $\mathbf{a}_t^r=\pi_\theta^r(\mathbf{q}_t,\mathbf{z}_t^f,\mathbf{g}_t \cdot \mathbf{z}_t^\tau,\mathbf{a}_t^b, \mathbf{a}_{t:t+K}^b)$, where $\mathbf{q}_t$ is the proprioception, $\mathbf{z}_t^f$ is the flow representation, $\mathbf{z}_t^\tau$ is the tactile representation, $\mathbf{g}_t$ is a contact gate to add tactile as input, and $\mathbf{a}_{t:t+K}^b$ is the base policy action chunk. For training details, we keep $\pi_{\text{b}}$ frozen, learn a residual tactile policy $\pi_\theta^r$, and keep optimizing the flow-tactile critic $Q_{\eta}$. 

During execution, we keep the current keypoint condition $\hat{\mathbf{c}}_\ell$ fixed until the observed object keypoints are close enough, which enables close-loop corrections with the keypoints as subgoal. Following PLD~\cite{pld}, we add residual scale with a scheduler to ensure safe exploration. 
 
\vspace{-10pt}
\subsection{Multi-Sensory Reward Design}
\label{sec:reward}
\vspace{-6pt}

Real-world RL is often limited by the difficulty of designing dense rewards. To guide the robot toward the near-contact states and encourage safe contact, our pipeline uses a normalized multi-sensory reward that combines generated object-centric flow with tactile  signal: $
r_t = \mathrm{normalize}(w_r r_t^{\text{reach}} + w_g r_t^{\text{grasp}} + w_f r_t^{\text{flow}}) - w_s r_t^{\text{safety}}.$
The generated flow provides task-level, object-centric motion guidance, while tactile feedback encourages stable and safe exploration. When the task succeeds, the operator assigns a terminal success reward of $1$. Before grasping, the reaching reward $r_t^{\text{reach}}$ guides the end-effector toward the initial 3D object centroid $\mathbf{c}_0^{3D}$; after grasping, the flow reward $r_t^{\text{flow}}$ tracks sparse subgoals from the generated object flow generated from Sec.~\ref{sec:base}~\cite{genflowrl,dex4d}:
\[
\vspace{-2pt}
\small
\begin{gathered}
r_t^{\text{reach}} =
1-\mathrm{clip}\!\left(
\frac{\|\mathbf{p}_t^{ee}-\mathbf{c}_0^{3D}\|_2}{d_{\max}^{\text{reach}}},0,1
\right),
\quad
d_t^{\text{flow}} =
\left\|
\mathbf{T}_{t_0\rightarrow t}^{\text{rel}}
-
\hat{\mathbf{T}}_{t_0\rightarrow \ell}^{\text{rel}}
\right\|_2, \\
r_t^{\text{flow}} =
\frac{\ell_t}{L}
+
\frac{1}{L}
\left(
1-\mathrm{clip}\!\left(
\frac{d_t^{\text{flow}}}{d_{\max}^{\text{flow}}},0,1
\right)
\right).
\end{gathered}
\]
Here, $\mathbf{p}_t^{ee}$ is the end-effector position, $\mathbf{c}_0^{3D}$ is the initial 3D object centroid, $d_{\max}^{\text{reach}}$ and $d_{\max}^{\text{flow}}$ are normalization constants, $\mathbf{T}$ represent the observed object motion and the generated subgoal motion, $d_t^{\text{flow}}$ is their transformation error, $L$ is the total number of sparse subgoals, and $\ell_t\in\{0,\dots,L \}$ is the number of subgoals reached by time $t$.
When $d_t^{\text{flow}}<\epsilon_{\text{flow}}$, the current subgoal is considered reached and the policy switches to the next sparse subgoal.

We further add tactile rewards to encourage stable contact and penalize unsafe interaction:
\vspace{-3pt}
\[
\small
r_t^{\text{grasp}}
=
\mathbf{1}\!\left[
\frac{1}{|\Omega|}
\sum_{u\in\Omega}\mathbf{D}_t^\tau(u)
>
\epsilon_{\text{depth}}
\right],
\qquad
r_t^{\text{safety}}
=
\mathbf{1}[m_t>\epsilon_{\text{safety}}].
\]
Here, $\mathbf{D}_t^\tau$ is the tactile depth image, $\Omega$ is the valid tactile region, $u$ indexes tactile pixels, $\epsilon_{\text{depth}}$ is the contact threshold, $m_t$ is the average tactile marker displacement, and $\epsilon_{\text{safety}}$ is the safety threshold. If the safety penalty is triggered, we assign a large negative reward and reset the robot.
 
\vspace{-10pt}
\section{Experiments}
\vspace{-6pt}
 
\begin{figure*}[t]
    \centering

    \includegraphics[width=\linewidth]{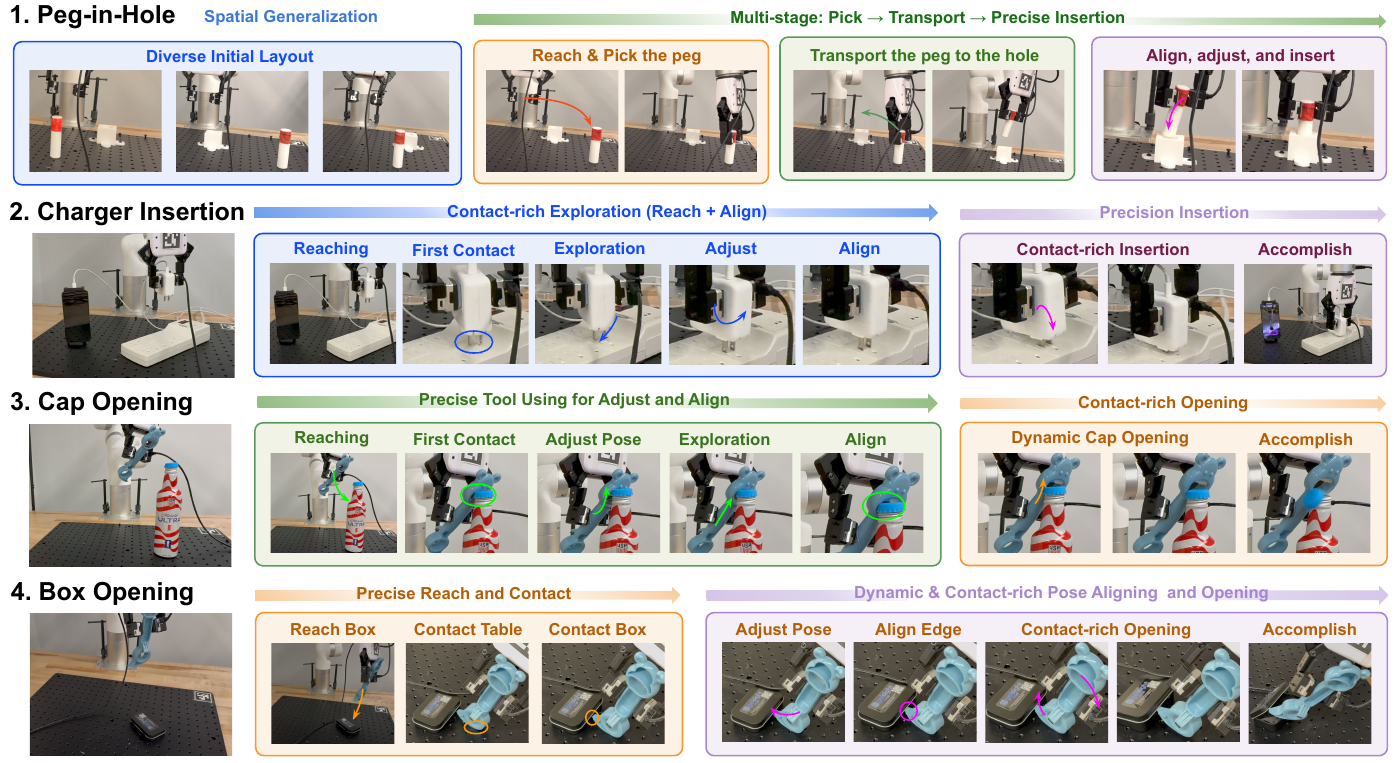}
    \caption{\textbf{Real-world contact-rich manipulation tasks.}
    We evaluate \ourframework{} on four tasks: \emph{peg-in-hole} requires spatial generalization and multi-stage insertion; \emph{charger insertion} requires contact exploration and precise insertion; \emph{cap opening} requires dynamic contact reasoning and pose adjustment; and \emph{box opening} requires precise edge alignment and contact-rich opening.}
    \label{fig:demos}

    \vspace{0.4em}

    \captionsetup{type=table}
    \caption{Final success rates across four real-world contact-rich manipulation tasks after RL training.}
    \label{tab:final_performance}
    \vspace{0.1cm}

    \resizebox{\textwidth}{!}{
    \begin{tabular}{lccccc}
        \toprule
        \textbf{Method} & \textbf{Peg-in-Hole} & \textbf{Charger Insertion} & \textbf{Cap Opening} & \textbf{Box Opening} & \textbf{Average} \\
        \midrule
        PLD*~\cite{pld} & 65\% & 60\% & 50\% & 35\% & 52.5\% \\
        PLD~\cite{pld} (Visual Only) & 60\% & 30\% & 40\% & 20\% & 37.5\% \\
        ViTAL~\cite{touchbegin} & 50\% & 50\% & 45\% & 30\% & 43.75\% \\
        Ours & \textbf{100\%} & \textbf{100\%} & \textbf{90\%} & \textbf{85\%} & \textbf{93.75\%} \\
        \bottomrule
    \end{tabular}
    }

    \captionsetup{type=figure}

    \vspace{0.4cm}

    \includegraphics[width=\textwidth]{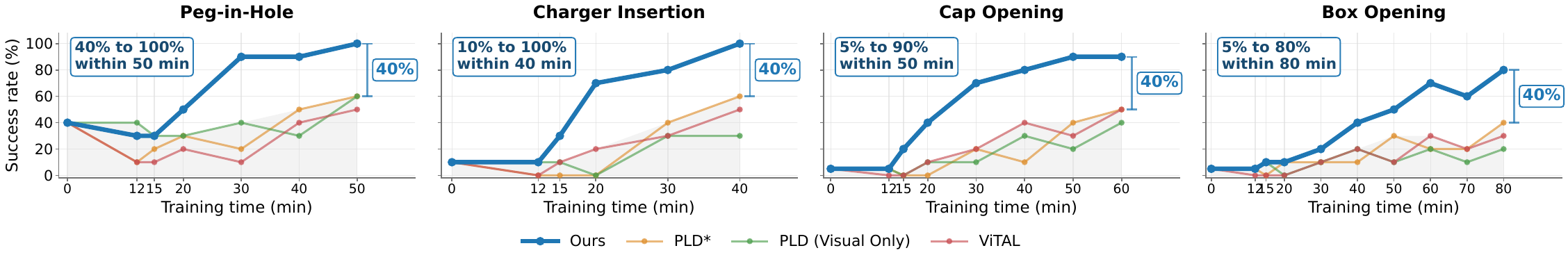}
    \vspace{-0.5cm}
    \caption{Comparison of RL training results across four different contact-rich manipulation tasks.}
    \label{fig:main}
    \vspace{-0.5cm}
\end{figure*}
In this section, we evaluate \ourframework{} along four axes: real-world learning efficiency for adapting tactile feedback (Sec.~\ref{sec:adapt}), compatibility with various visual base policies (Sec.~\ref{sec:general}), comparison to other visuo-tactile policy learning methods (Sec.~\ref{sec:il}), and generalizability across different tactile representations (Sec.~\ref{sec:tact}). Furthermore, we conduct extensive ablation studies on reward shaping, residual policy design, warm-start strategies, and action designs. We also conduct failure case evaluations and analysis of generated flows and base policies. See Appendix~\ref{app:experiment} for details.

\vspace{-8pt}

\paragraph{Tasks.}
We use an xArm7 with a gripper mounted with GelSight Mini and a calibrated Intel RealSense D435 third-view camera. We evaluate \ourframework on four challenging contact-rich manipulation tasks, shown in Fig.~\ref{fig:demos}, including: i) \textbf{Peg-in-Hole}. Reach and grasp a cylinder and insert it into a target receptacle. The task requires spatial generalizability across different locations, stable grasping during transport, and millimeter contact alignment during insertion; ii) \textbf{Charger Insertion}. Insert the charger into a power strip. This task is particularly challenging because of tiny charger head and the small port tolerance; iii) \textbf{Cap Opening}. A tool using task, where the robot use a cap opener to open a bottle cap. This task is challenging because it requires maintaining a precise opener pose while handling dynamic contact-rich interactions with the cap; and iv) \textbf{Box Opening}. A tool using task, where the robot use a lever-type bottle opener to open the GelSight box. This task is challenging because the opener must make precise contact with a very small cap edge and maintain contact-rich interaction while executing a highly dynamic levering motion. All those tasks are contact-rich manipulation tasks with dynamic and precise contact need tactile feedback.
 
 

 
\vspace{-8pt}
\subsection{Adapting Tactile Feedback with Residual RL}
\vspace{-6pt}\label{sec:adapt}
In this section, we evaluate \ourframework across all tasks and compare it with other real-world RL methods, which are adapted into our tactile adaptation setting. For each task, we collect 50 human demos using our data collection system in Sec.~\ref{sec:data} and train the human flow policy (in Sec.~\ref{sec:base}) as the base policy. Evaluated in 20 trials, the success rates of the base policies are 40\%, 10\%, 5\%, and 5\% for \textbf{Peg-in-Hole}, \textbf{Charger Insertion}, \textbf{Cap Opening}, and \textbf{Box Opening}, respectively. 

We implement three different baselines: i) \textbf{PLD*}~\cite{pld}: adapting PLD to our flow-tactile setting, collecting real-world data with tactile into the replay buffer, but no initialization and bootstrapping for the critic; ii) \textbf{PLD (visual only)}: as the original version of PLD, we initialize the critic for the flow policy via offline data, warm-start the replay buffer, and do the residual RL; iii) \textbf{ViTAL}~\cite{touchbegin}: we train the residual policy and critic from scratch without warm-start. For each task, we perform a 12-minute warm-start phase for all methods except ViTAL, followed by online policy learning. The total training time is 50, 40, 60, and 80 minutes across the four tasks, respectively. We evaluate the policies in 10 trials for each checkpoint shown in the figure, and 20 trials for the final model.

As shown in Tab.~\ref{tab:final_performance} and Fig.~\ref{fig:main}, our method improves the weak human-flow base policies from only 5--40\% success rate to 100\%, 100\%, 90\%, and 85\% final success rate within 40--80 mins, outperforming all baselines with more than 40\% success rate. In contrast, \textbf{PLD*} improves more slowly and unstably, showing the importance of bootstrapping the tactile encoder and critic. \textbf{PLD (visual only)} performs better in the early stage due to its critic initialization, but achieves the worst overall results without adapting tactile feedback. \textbf{ViTAL} is less sample-efficient because it learns visuo-tactile critics from scratch. Overall, these results show that replay-buffer warm-start, tactile encoder and critic optimization, and tactile residual learning are all critical for efficient adaptation.

\vspace{-8pt}
\subsection{Adapting Tactile Feedback to Various Base Visual Policies}
\begin{figure*}[h]
    \vspace{-5pt}
    \centering
    \begin{minipage}[c]{0.52\textwidth}
    As shown in Fig.~\ref{fig:base}, \ourframework{} improves all base policies from 15--50\% to 75--100\% after 50 minutes of real-world practice. These results show that tactile residual adaptation is broadly compatible with different visual motion priors, rather than being tied to a specific policy architecture. Among them, the human flow policy achieves the highest performance, suggesting that human demos provide a smoother motion prior for contact-rich refinement. We provide further analysis of data and policy smoothness in the Appendix~\ref{app:experiment}, which helps explain why teleoperation data becomes low-quality and impractical for contact-rich manipulations.
    \end{minipage}
    \hfill
    \begin{minipage}[c]{0.45\textwidth}
    \vspace{-0.2cm}
        \centering
        \includegraphics[width=\linewidth]{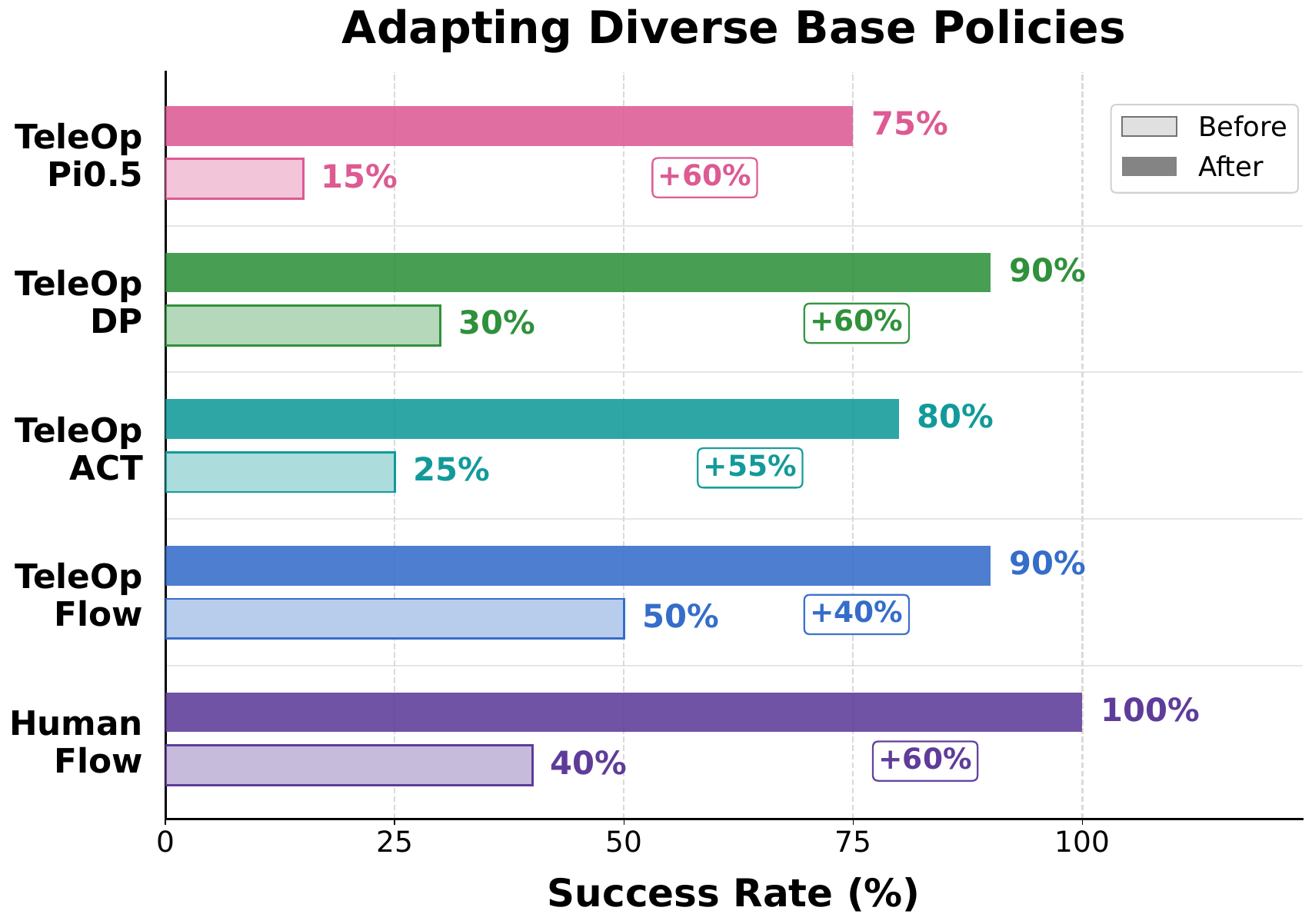}
        \caption{\ourframework consistently improves diverse base visual policies.}
        \label{fig:base}
    \end{minipage}
    \vspace{-0.35cm}
\end{figure*}

\vspace{-6pt}\label{sec:general}
In this section, we evaluate \ourframework on the \textbf{Peg-in-Hole} task across different base policies trained from different data sources. We evaluate it on five different base policies: i) Human Flow Policy in Sec.~\ref{sec:base}, with initial success rate 40\%; ii) Teleoperation-based Flow Policy, with initial success rate 50\%; iii) Teleoperation-based \texttt{ACT~\cite{act}}, with initial success rate 20\%; iv) Teleoperation-based \texttt{DP~\cite{dp}}, with initial success rate 30\%; v) fine-tuned \texttt{$\pi_{0.5}$~\cite{pi05}}, with initial success rate 20\%. In this section, we only evaluate the performance on the \textbf{Peg-in-Hole} task because collecting high-quality teleoperation data for the other tasks is challenging due to the contact-rich interactions, making the demonstrations too noisy to policy training. The result is shown in Fig.~\ref{fig:base}.

\vspace{-8pt}
\subsection{Comparison with Other Visuo-Tactile Robot Learning Methods}
\vspace{-6pt}
\label{sec:il}
To evaluate the data efficiency of \ourframework, we compare against visuo-tactile policy learning pipelines built on different policy backbones: \texttt{ACT}, \texttt{DP}, and \texttt{$\pi_{0.5}$}. For \texttt{ACT}, we adopt a straightforward visuo-tactile fusion strategy by concatenating visual and tactile features. For \texttt{DP}, we implement \texttt{RDP~\cite{rdp}}, a SOTA diffusion-based visuo-tactile learning method. For \texttt{$\pi_{0.5}$}, we incorporate additional tactile tokens and conduct supervised fine-tuning~\cite{flexitac}. To ensure a fair comparison, all baselines are trained with additional teleoperation data collected under the same time budget as our online training stage (i.e., 50 minutes), increasing the number of demonstrations from 50 to 90. 

\begin{table}[h]
\vspace{-8pt}
\centering
\caption{Success rates of different visuo-tactile robot learning methods on the \textbf{Peg-in-Hole} task.}
\label{tab:peg}
\resizebox{\linewidth}{!}{
\begin{tabular}{l
>{\columncolor{green!15}}c
>{\columncolor{green!15}}c
>{\columncolor{blue!15}}c
>{\columncolor{blue!15}}c
>{\columncolor{yellow!20}}c
>{\columncolor{yellow!20}}c
>{\columncolor{green!15}}c
>{\columncolor{blue!15}}c
>{\columncolor{yellow!20}}c
c}
\toprule
Task & ACT~\cite{act} & ACT + Tactile & DP~\cite{dp} & RDP~\cite{rdp} & $\pi_{0.5}$~\cite{pi05} & $\pi_{0.5}$ + Tactile & \textbf{Ours (ACT)} & \textbf{Ours (DP)} & \textbf{Ours ($\pi_{0.5}$)} & \textbf{Ours} \\
\midrule
\textbf{Peg-in-Hole} & 25\% & 60\% & 30\% & 65\% & 15\% & 45\% & \textbf{80\%} & \textbf{90\%} & \textbf{75\%} & \textbf{100\%} \\
\bottomrule
\end{tabular}
}
\vspace{-4pt}
\end{table}

As shown in Tab.~\ref{tab:peg}, imitation-learning-based visuo-tactile policy learning remains less effective than \ourframework, which achieves 20\%--30\% higher success rates for all backbones and reaches 100\% success with our human flow policy. This  shows online tactile residual refinement is more effective than learning a visuo-tactile policy from extra teleoperated demonstrations, highlighting that tactile correction for contact-rich manipulation is better learned through trial-and-error.

\vspace{-8pt}
\subsection{Adaptation Across Different Tactile Representations}
\vspace{-6pt}\label{sec:tact}
To highlight its compatibility, we further evaluate \ourframework across different tactile representations in two different contact-rich manipulation tasks, \textbf{Peg-in-Hole} and \textbf{Charger Insertion}. We compare  three pretrained tactile image encoders, \texttt{AnyTouch2~\cite{anytouch2}}, \texttt{Sparsh~\cite{sparsh}}, and \texttt{T3~\cite{t3}}, as well as low-dimensional \texttt{Tactile Markers} with two-layer MLP. For pretrained tactile encoders, we freeze the backbone and only optimize the final projection layers and a lightweight adaptive layer that aligns with the flow features. For markers, we optimize their encoder from scratch.

\vspace{-6pt}
\begin{figure*}[h]
    \centering
    \begin{minipage}[c]{0.57\textwidth}
        As shown in Fig.~\ref{fig:tactile}, our method consistently improves performance across all tactile representations, which demonstrates that our method is not tied to a particular setting. On both \textbf{Peg-in-Hole} and \textbf{Charger Insertion} tasks, \texttt{AnyTouch2} and low-dimensional tactile markers achieve comparable final performance, which highlights that \ourframework can effectively adapt with both pretrained tactile features and compact marker-based tactile signals. Meanwhile, we observe that \texttt{T3} and \texttt{Sparsh} perform worse than \texttt{AnyTouch2} on Charger Insertion, where the task involves more dynamic contact. The potential reason is that \texttt{T3} and \texttt{Sparsh} are not pretrained on dynamic manipulation tactile data.
    \end{minipage}
    \hfill
    \begin{minipage}[c]{0.4\textwidth}
    \vspace{-0.1cm}
        \centering
        \includegraphics[width=\linewidth]{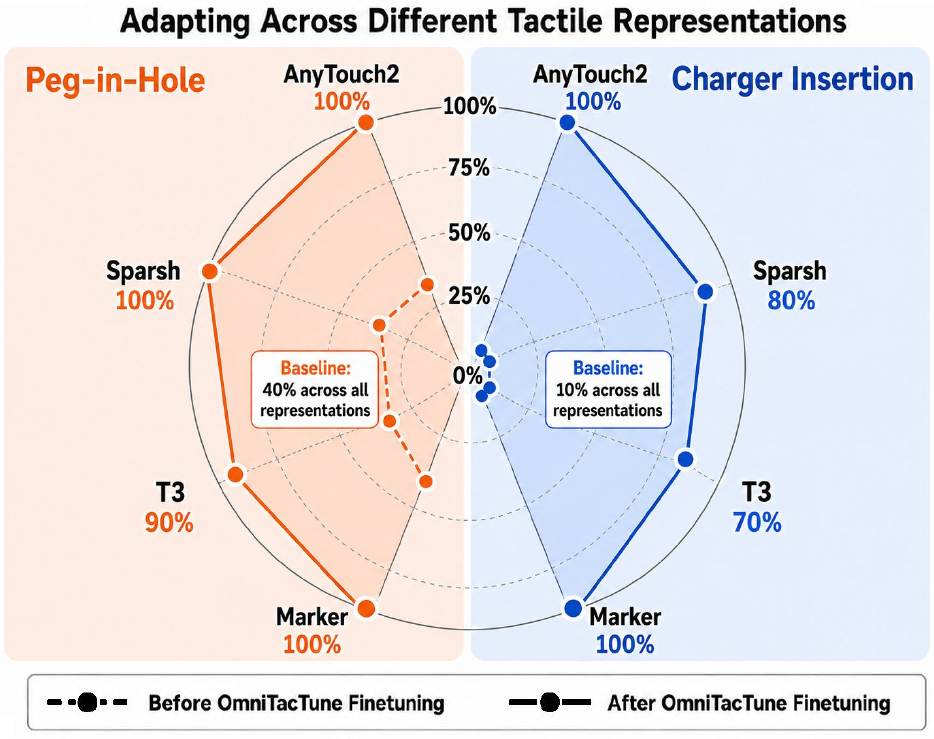}
            \caption{Adapting different tactile representations via \ourframework.}
        \label{fig:tactile}
    \end{minipage}
\vspace{-12pt}
\end{figure*}
\vspace{-4pt}

\section{Conclusion and Limitation}
\vspace{-5pt}
We presented \ourframework, a policy-agnostic real-world RL pipeline that adapts tactile feedback to pretrained visual policies through residual correction. \ourframework enables efficient tactile adaptation without offline tactile demonstrations by using visual base policies as motion priors, warm-starting tactile-aware critics and encoders from autonomous rollouts, and incorporating multi-sensory reward shaping. Across four challenging real-world contact-rich manipulation tasks, our method improves visual base policies from 5--40\% success to 85--100\% within 40--80 minutes, while generalizing across tasks, base policies, and tactile representations. 

\textbf{Limitation.} Despite these results, \ourframework still inherits common limitations of real-world RL, including the need for manual resets and hardware wear under repeated contact-rich interactions, especially when using fragile vision-based tactile sensors. Future work should reduce reset and hardware costs through safer and more automated real-world training, while further improving motion priors and data augmentation, such as incorporating world models~\cite{vlaw,unisim,dreamerv3,letacmpc} to generate more physically plausible visuo-tactile rollouts for pretraining and online policy improvement. Moreover, adapting it into other embodiments and tactile sensor is a straightforward extension.
 


\clearpage
 
\bibliography{references}  
\newpage

\appendix
\section*{Appendix}

\section{Implementation Details}
\label{app:implementation}

\subsection*{Data Collection System}
\label{app:data_collection}
\begin{figure*}[!h]
    \centering
    \includegraphics[width=0.7\textwidth]{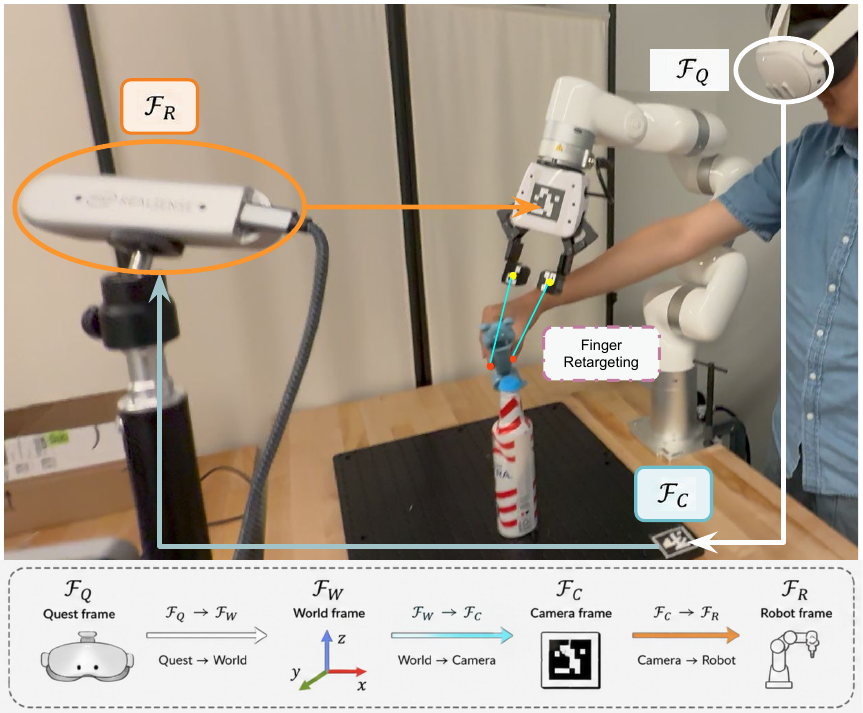}
    \caption{Visualization of our data collection system, for both hand and finger retargeting.}
    \label{fig:retarget}
    \vspace{-10pt}
\end{figure*}
\paragraph{Human video collection and Quest pose retargeting.}
For each task, we collect human videos using a third-view RGB camera and a Meta Quest headset. 
The RGB camera records the in-scene videos for human hand operations, while the Quest headset provides the 6-DoF hand pose and the finger keypoints in the Quest frame. Following prior VR-based data collection systems~\cite{openteach}, we calibrate the Quest frame, the world frame, and the robot base frame via an ArUco marker placed on the table. 
Let ${}^{q}\mathbf{T}_{h}(t)$ denote the tracked hand pose in the Quest frame at time $t$, ${}^{w}\mathbf{T}_{q}$ show the transform from the Quest frame to the world frame, and ${}^{r}\mathbf{T}_{w}$ denote the transform from the world frame to the robot base. 
The robot-aligned hand pose is obtained by
\begin{equation}
    {}^{r}\mathbf{T}_{h}(t)
    =
    {}^{r}\mathbf{T}_{w}
    {}^{w}\mathbf{T}_{q}
    {}^{q}\mathbf{T}_{h}(t).
\end{equation}
We synchronize the third-view video and Quest pose stream, and linearly interpolate the lower-frequency stream from the quest. To reduce tracking noise, we apply a small smoothing filter to the tracked hand translation and use spherical interpolation for rotations.

\paragraph{Finger pose tracking and retargeting.}
For demonstrations that require precise grasp, we further use the Quest hand skeleton to estimate finger motion and retarget them to the binary grasping signal. Because human hand's skeleton and the robot gripper have different morphology and degrees of freedom, we do not directly copy human joint angles. Instead, we retarget the finger geometry into a compact binary gripper command. Specifically, we extract the thumb and index fingertip positions, their relative distance, and the palm normal direction from the human hand skeleton. the finger distance is computed from the normalized thumb-index distance:
\begin{equation}
    a^{\mathrm{grip}}_t
    =
    \mathrm{clip}
    \left(
    \alpha
    \frac{
    \| \mathbf{p}^{\mathrm{thumb}}_t - \mathbf{p}^{\mathrm{index}}_t \|_2
    - d_{\min}}
    {d_{\max}-d_{\min}}
    + \beta,\,
    a_{\min}, a_{\max}
    \right),
\end{equation}
where $\mathbf{p}^{\mathrm{thumb}}_t$ and $\mathbf{p}^{\mathrm{index}}_t$ are the tracked fingertip positions, $d_{\min}$ and $d_{\max}$ are calibration distances for closed and open hand poses, and $\alpha,\beta$ map the human aperture to the robot gripper range. 
We further detect grasp intent by thresholding the temporal change of the thumb-index distance and the absolute aperture value, which produces a binary grasp signal used for grasping and placing.

\paragraph{Teleoperation data collection.}
In addition to human videos, we collect robot demonstrations through Meta Quest headset via an OpenTeach-style interface~\cite{openteach}. 
The operator controls the robot end-effector with hand motion, while the system records synchronized RGB observations, robot proprioception, gripper state, and executed delta actions. We use the same calibrated third-view RGB camera and robot setup as in the human-video demonstrations. 
Each teleoperated episode is stored as
$\{I_t, \mathbf{q}_t, \mathbf{a}_t\}_{t=1}^{T}$,
where $I_t$ is the third-view RGB observation, $\mathbf{q}_t$ is the robot proprioceptive state, and $\mathbf{a}_t$ is the executed delta Cartesian end-effector action with gripper command. 
Teleoperation data is mainly used to train robot-data-based base policies, including ACT, Diffusion Policy, and $\pi_{0.5}$. 
For contact-rich tasks like \textbf{Charger Insertion}, \textbf{Box Opening}, and \textbf{Cap Opening}, it's impractical to collect high-quality during the contact stage because human cannot get the contact event with their hand~\cite{mimictouch}. Therefore, in our main results, we use human flow policies for all four tasks and use teleoperation policies only for the \textbf{Peg-in-Hole}. More analysis has been shown in Appendix.~\ref{app:analysis}

\subsection{Flow Generator}
\label{app:flow_generator}

\begin{figure*}[!h]
    \centering
    \begin{minipage}[c]{0.38\textwidth}
        \small
        Our flow generator follows the object-centric design of Im2Flow2Act and GenFlowRL~\cite{im2flow2act,genflowrl}. 
        Given the first RGB frame $I_0$, we first localize the manipulated object using DINOv2 features and prompt SAM to obtain an object mask $M_0$. 
        We then sample $N=32$ keypoints inside the object mask and track them through the demonstration video using CoTracker 3~\cite{cotracker3}. 
        This generates an object keypoint trajectory
        \begin{equation}
        \mathbf{K}_{1:T}=\{\mathbf{k}_t^i\}_{t=1,i=1}^{T,N}.
        \end{equation}

        The flow generator $G_{\phi}$ predicts a task-specific object flow from the initial image, keypoints, and task condition:
        \begin{equation}
            \hat{\mathbf{F}}
            =
            G_{\phi}(I_0, \mathbf{K}_0, z),
        \end{equation}
        where $\mathbf{K}_0$ is the initial set of object keypoints and $z$ denotes the task condition. 
        We initialize $G_{\phi}$ from the pretrained flow generator of Im2Flow2Act and fine-tune it on our collected demonstrations. 
        For each task, we fine-tune the generator with 50 demonstrations using LoRA adapters~\cite{lora} while keeping the main pretrained  
    \end{minipage}
    \hfill
    \begin{minipage}[c]{0.6\textwidth}
        \centering
        \includegraphics[width=\linewidth]{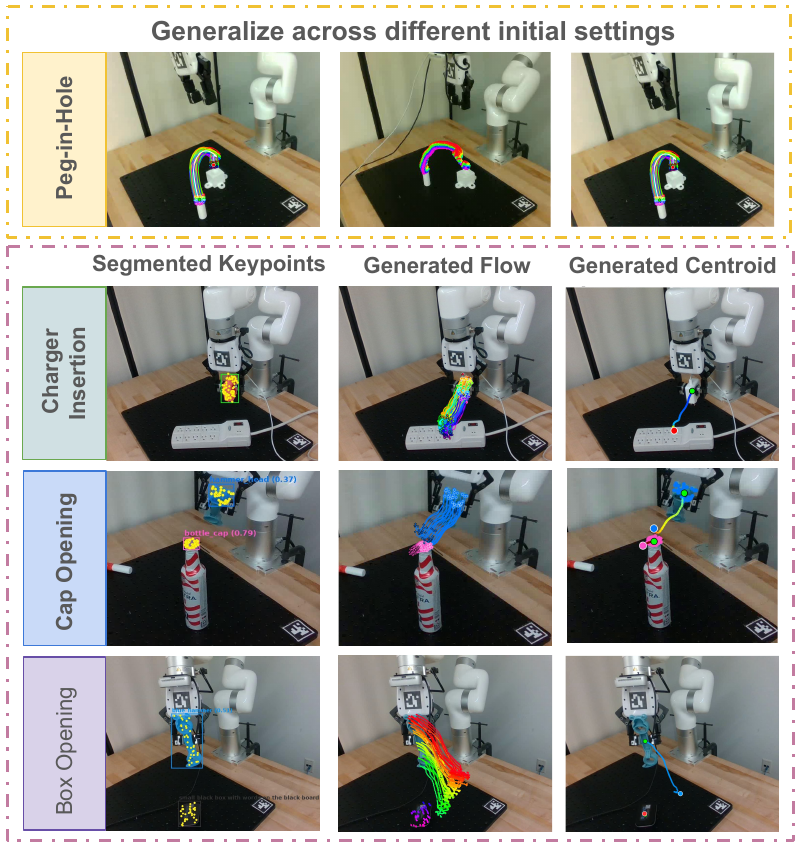}
        \caption{Generated flows for each tasks.}
        \label{fig:flow}
    \end{minipage}
    \vspace{-10pt}
\end{figure*}
backbone frozen. This improves task-specific flow generation while avoiding overfitting to the small demonstration set.

The generator is trained with a point-wise flow prediction loss and a temporal smoothness regularizer:
\begin{equation}
    \mathcal{L}_{\mathrm{flow}}
    =
    \frac{1}{TN}
    \sum_{t=1}^{T}
    \sum_{i=1}^{N}
    \left\|
    \hat{\mathbf{k}}^i_t - \mathbf{k}^i_t
    \right\|_2^2
    +
    \lambda_{\mathrm{smooth}}
    \sum_{t=2}^{T-1}
    \left\|
    \hat{\mathbf{k}}_{t+1}
    -2\hat{\mathbf{k}}_{t}
    +\hat{\mathbf{k}}_{t-1}
    \right\|_2^2 .
\end{equation}
At test time, the generator predicts a complete object-centric flow from the initial scene. To formulate our flow policy and the object-centric reward, we sparsify the generated trajectory into $L$ subgoals and use these subgoals as compact guidance for the flow-based base policy and dense object-centric reward during RL. This follows from the observation that object flow provides a cross-embodiment manipulation interface with less embodiment gap.

\subsection{Base Visual Policies}
\label{app:base_policies}

\paragraph{Flow policy.}
The flow policy converts generated and observed object flows into compact object-centric policy inputs. 
At time $t$, we compute the current object centroid $\mathbf{c}_t$, the relative transformation between current object centroid and the initial object centroid$\mathbf{T}^{\mathrm{rel}}_{t_0\rightarrow t}$, the selected generated lookahead centroid $\hat{\mathbf{c}}_{\ell}$, the  relative transformation between the generated centroid and the initial centroid$\hat{\mathbf{T}}^{\mathrm{rel}}_{t_0\rightarrow \ell}$, and the remaining transformation between the current centroid and the generated lookahead centroid $\mathbf{T}^{\mathrm{rel}}_{t\rightarrow \ell}$. 
The policy input is
\begin{equation}
    \mathbf{o}^{\mathrm{flow}}_t
    =
    [
    \mathbf{c}^{3D}_0,\,
    \mathbf{c}_t,\,
    \mathbf{T}^{\mathrm{rel}}_{t_0\rightarrow t},\,
    \hat{\mathbf{c}}_{\ell},\,
    \hat{\mathbf{T}}^{\mathrm{rel}}_{t_0\rightarrow \ell},\,
    \mathbf{T}^{\mathrm{rel}}_{t\rightarrow \ell}
    ],
\end{equation}
where $\mathbf{c}^{3D}_0$ is the initial 3D object centroid used for reaching and grasping. 
The flow policy predicts an action chunk:
\begin{equation}
    \mathbf{a}^{\mathrm{b}}_{t:t+K}
    =
    \pi^{\mathrm{flow}}_{\theta}
    (
    \mathbf{o}^{\mathrm{flow}}_t,
    \mathbf{q}_t
    ),
\end{equation}
where $\mathbf{q}_t$ is robot proprioception and $K$ is the action horizon. 
Unlike diffusion-based visuomotor policies, we implement the flow policy as a lightweight action-chunking policy over low-dimensional object-centric states. In practice, all modules are small MLPs with LayerNorm. This design keeps inference fast and stable during online RL, while they can follow the generated motion priors learned from expert demonstrations. 
During execution, the current lookahead subgoal $\ell$ is kept fixed until the observed object transformation is close to the generated subgoal, and then the policy switches to the next sparse subgoal along the generated flow. We sparsify the generated trajectory into $L$ subgoals, which enable the robot to reach the goal in a closed-loop manner. This closed-loop subgoal tracking is similar in spirit to point-track-conditioned policies such as Dex4D~\cite{dex4d}.

\paragraph{Diffusion Policy.}
We train Diffusion Policy (DP)~\cite{dp} on teleoperated robot demonstrations. The policy takes third-view RGB observations and robot proprioception as input and predicts a sequence of future end-effector actions through conditional denoising diffusion. We use the same image encoder, action horizon, and observation horizon as the prior work across all DP-based comparisons. Because DP can produce smoother action chunks than one-step behavior cloning, it provides a strong visual base policy for contact-rich tasks. However, it still open-loop for short-horizon motions and lacks direct contact feedback, so it may fail during the contact-rich manipulation tasks and require the close-loop corrections.

\paragraph{ACT.}
We train Action Chunking Transformer (ACT)~\cite{aloha2} on the same teleoperation dataset. ACT encodes RGB observations and proprioception with a transformer-based conditional VAE and predicts an action chunk instead of a single action. We use the standard ACT training objective, including the reconstruction loss for actions and the KL regularization term for the latent variable. At test time, we use temporal ensembling over overlapping action chunks to reduce action jitter. Same as DP, although it can predict smooth action chunks, ACT still lacks closed-loop control with contact reasoning, making it a useful stress test for our residual tactile adaptation pipeline.

\paragraph{\textbf{$\pi_{0.5}$.}}
For VLA experiments, we fine-tune $\pi_{0.5}$~\cite{pi05} on our teleoperation demonstrations. The model takes RGB observations from a single camera and a language instruction describing the task, such as ``insert the peg into the hole'' or ``insert the charger into the port'', and predicts low-level robot actions. We use parameter-efficient fine-tuning via LoRA~\cite{lora} for the in our \textbf{Peg-in-Hole} task and keep the visual-language backbone fixed. Notably, since all the other settings only using a single $3$-rd view RGB camera, $\pi_{0.5}$ also uses only one camera.

\subsection{Baselines}
\label{app:baselines}

\paragraph{PLD*.}
We adapt PLD~\cite{pld} to our setting by using the frozen visual base policy as the generalist policy and training a residual actor with real-world interaction. 
Different from its original setting, the residual policy receives the same visual/flow + tactile observations as our method and is trained with SAC. Different from the original PLD, since the expert demo doesn't have tactile input, we cannot initialize the critic with the offline data. During the warm-start stage, we only autonomous rollouts the base policy to restore the replay buffer with the random critic. It therefore tests whether residual RL alone is sufficient without our fine-tuning our the tactile encoder and initialize the critic.

\paragraph{PLD (Visual Only).}
This baseline follows the visual-only residual RL setting more closely. We initialize the critic using the available offline visual demonstrations and train a residual policy on top of the flow base policy. The policy does not use tactile observations, tactile rewards, or tactile encoder optimization. This baseline tests whether the improvement comes from residual RL on the visual policy alone or from adapting tactile feedback.

\paragraph{ViTAL.}
We implement ViTAL as a visuo-tactile real-world RL baseline that learns a visuo-tactile policy and critic from scratch without any critic initialization, replay buffer initialization, and encoder optimization. The policy receives flow observations, proprioception, and tactile observations, and is trained using the same online interaction budget. This baseline evaluates the importance of using warm-start stage in critic initialization and encoder optimization.

\paragraph{ACT + Tactile.}
For the ACT tactile baseline, we simply add tactile observations to the ACT policy through feature concatenation. The RGB image is encoded by the original visual encoder, and the tactile image is encoded by the AnyTouch2~\cite{anytouch2} used in our method. The two features are concatenated with proprioception and passed to the ACT transformer and predict action chunks. This baseline is trained by  imitation learning from the teleoperated demonstrations with tactile observations. 
Unlike our method, it requires paired visuo-tactile demonstrations and does not improve through online tactile practice.

\paragraph{RDP.}
For the diffusion-based tactile baseline, we implement Reactive Diffusion Policy (RDP)~\cite{rdp}, a slow-fast visuo-tactile policy for contact-rich manipulation. The slow branch predicts a visual motion plan from RGB observations. Then, the fast branch encode the latent plan from the slow policy with the tactile feedback, reacts to the contact and refines the action during execution. We train RDP on the same amount of visuo-tactile teleoperation data used by the other visuo-tactile imitation baselines. 
This baseline tests whether a SOTA visuo-tactile diffusion policy can match the data-efficiency of our online tactile residual adaptation or not.

\paragraph{$\pi_{0.5}$ + Tactile.}
For the VLA tactile baseline, we follow recent tactile-augmented VLA designs such as LeFlexiTac~\cite{flexitac}. We encode tactile observations as additional tactile tokens and append them to the visual-language-action input during supervised fine-tuning. The model is fine-tuned on visuo-tactile teleoperation data with the same data budget as the other imitation baselines. Compared with our residual adaptation, this baseline directly fine-tune the VLA policy through supervised fine-tuning via LoRA~\cite{lora}, and therefore requires paired teleoperated, visual-tactile demonstrations for each task.

\subsection{Warm-start Implementation}
\label{app:warmstart}

During warm-start, we execute the frozen base policy in the real world and set the residual action to zero:
\begin{equation}
    \mathbf{a}_t = \mathbf{a}^{\mathrm{b}}_t,
    \qquad
    \mathbf{a}^{\mathrm{r}}_t = \mathbf{0}.
\end{equation}
The goal of this stage is not to improve the policy immediately, but to collect on-policy tactile experience under the deployment distribution of the base policy. As shown in ~\cite{effrl}, without initializing the replay buffer and bootstrapping the critic is easily to lead the policy forget. In our setting, since the expert doesn't contains tactile feedback, we need to initialize the buffer and critic through online data during interaction.
During data collection, each transition is stored as
\begin{equation}
    (
    \mathbf{q}_t,\,
    \mathbf{z}^{f}_t,\,
    \mathbf{o}^{\tau}_{t-1:t},\,
    \mathbf{a}^{\mathrm{b}}_t,\,
    r_t,\,
    \mathbf{q}_{t+1},\,
    \mathbf{z}^{f}_{t+1},\,
    \mathbf{o}^{\tau}_{t+1-1:t+1}
    ),
\end{equation}
where $\mathbf{z}^{f}_t$ is the flow/object-centric representation, $\mathbf{o}^{\tau}_{t-1:t}$ is concatenated two tactile observations, and $r_t$ is the multi-sensory reward.

We initialize the tactile encoder from a pretrained tactile representation model, such as AnyTouch2~\cite{anytouch2}, Sparsh~\cite{sparsh}, or T3~\cite{t3}. 
For image-based tactile encoders, the backbone is frozen, and only the final projection layer and a lightweight adapter of the encoder are optimized. 
For marker-based tactile observations, we train a two-layer MLP encoder from scratch. 
The tactile representation is computed from consecutive tactile observations:
\begin{equation}
    \mathbf{z}^{\tau}_t
    =
    E_{\psi}
    (
    \mathbf{o}^{\tau}_{t-1:t}
    ).
\end{equation}
We only update the tactile encoder on contact transitions, where contact is detected by marker displacement or tactile depth thresholding. 
The flow-tactile critic follows the SAC formulation and is implemented as twin three-layer MLP critics. 
Each critic takes the robot state, flow representation, tactile representation, and executed action as input:
\begin{equation}
    Q_{\eta_i}
    (
    \mathbf{q}_t,
    \mathbf{z}^{f}_t,
    \mathbf{z}^{\tau}_t,
    \mathbf{a}_t
    ),
    \quad i \in \{1,2\}.
\end{equation}
During warm-start, since the residual action is set to zero, the executed action is given by the frozen base policy, i.e., $\mathbf{a}_t=\mathbf{a}^{\mathrm{b}}_t$. 
The twin critics are optimized with the Bellman loss:
\begin{equation}
    \mathcal{L}_{Q}
    =
    \sum_{i=1}^{2}
    \mathbb{E}_{\mathcal{B}}
    \left[
    \left(
    Q_{\eta_i}
    (
    \mathbf{q}_t,
    \mathbf{z}^{f}_t,
    \mathbf{z}^{\tau}_t,
    \mathbf{a}_t
    )
    -
    y_t
    \right)^2
    \right],
\end{equation}
where the target is computed as
\begin{equation}
    y_t
    =
    r_t
    +
    \gamma
    \min_{i=1,2}
    \bar{Q}_{\bar{\eta}_i}
    (
    \mathbf{q}_{t+1},
    \mathbf{z}^{f}_{t+1},
    \mathbf{z}^{\tau}_{t+1},
    \mathbf{a}_{t+1}
    ).
\end{equation}
Here, $\bar{Q}_{\bar{\eta}_i}$ denotes the target critic network, and $\mathbf{a}_{t+1}$ is also produced by the frozen base policy during warm-start. 
The tactile projection layer is trained with both critic supervision and a reconstruction regularizer:
\begin{equation}
    \mathcal{L}_{E}
    =
    \mathcal{L}_{Q}
    +
    \lambda_{\mathrm{rec}}
    \mathcal{L}_{\mathrm{rec}}.
\end{equation}
The reconstruction decoder is kept frozen, so only the tactile projection layer and lightweight adapter are updated. This warm-start stage bootstraps both the tactile representation and the tactile-aware critic before learning residual actions.

\paragraph{ControlTac trajectory-level augmentation.}~\label{app:controltac}
ControlTac~\cite{controltac} is a SOTA tactile data augmentation method. Because the warm-start buffer contains only a small number of real contact trajectories, we use it to increase tactile data diversity without additional robot interaction. ControlTac is a controllable tactile generation model that synthesizes realistic tactile images from a reference tactile observation, conditioned on desired contact-force changes and contact-pose variations. In our pipeline, it is used only to augment tactile observations using the contact force: the robot states, actions, flow features, rewards, and terminal labels are kept unchanged, while the tactile images are replaced by generated variants that simulate different but physically plausible contact conditions.

After each warm-start rollout, we first identify contact segments using the tactile marker displacement threshold. For each contact trajectory, we perform trajectory-level generation rather than frame-wise independent generation, so that the augmented tactile sequence remains temporally consistent across the rollout. Specifically, for each augmentation, we sample a force perturbation $\Delta f \in [-3, 10]$ and apply it consistently to all tactile samples in the same trajectory. We also apply domain randomization to each generated sample by sampling a pose/control offset from $\{-1,-0.5,0,0.5,1\}$, which introduces small variations in contact location and deformation. For every real trajectory, we generate two augmented tactile trajectories. 
The augmented transitions are then added to the replay buffer together with the real transitions and are used for tactile encoder adaptation and flow-tactile critic bootstrapping. This augmentation exposes the critic and tactile encoder to richer contact-force and contact-pose variations while preserving the same task-level motion and reward structure.

\begin{figure*}[!h]
    \centering
    \includegraphics[width=0.7\textwidth]{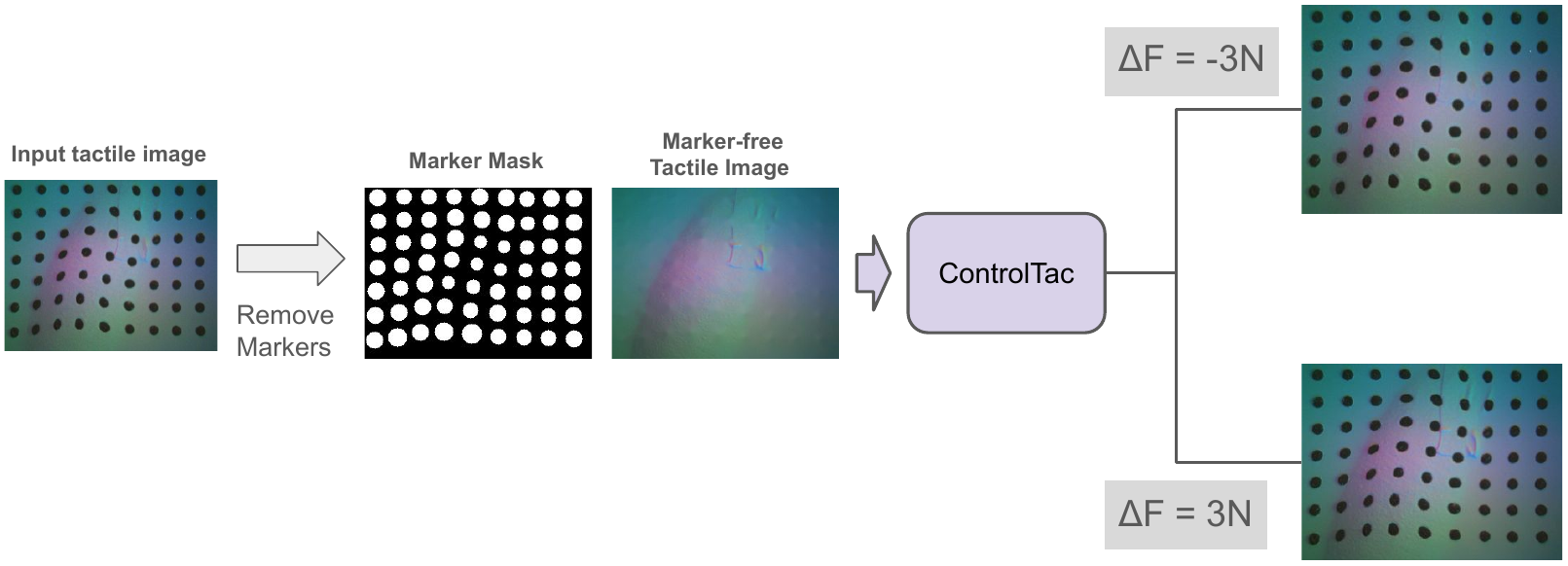}
    \caption{Visualization of ControlTac Generation Process.}
    \label{fig:control_aug}
    \vspace{-10pt}
\end{figure*}

For the implementation details of ControlTac, given a real tactile image, we first need to remove the visual marker pattern from the underlying tactile deformation because the training dataset of ControlTac is markless. Specifically, we detect the black marker locations and construct a binary marker mask, where marker regions are set to one and the remaining tactile surface is set to zero. We then remove the markers from the original tactile image through image inpainting, producing a marker-free tactile observation that preserves the contact-induced color and geometry changes. This marker-free image is used as the reference input to ControlTac, which generates new tactile images under different force conditions by perturbing the normal force, e.g., $\Delta F=-3$N and $\Delta F=+3$N. Since the generated outputs do not contain the original marker pattern, we compose the generated marker-free image with the original marker mask by copying the marker pixels back to their corresponding locations. The progress is visualized in Fig.~\ref{fig:control_aug}

\subsection{Online Residual Policy Learning}
\label{app:online_rl}

After warm-start, we keep the base policy frozen and train both a residual tactile policy and the critic via SAC. 
The final action is
\begin{equation}
    \mathbf{a}_t
    =
    \mathbf{a}^{\mathrm{b}}_t
    +
    \mathbf{s}_t\mathbf{a}^{\mathrm{r}}_t.
\end{equation}
The residual policy is conditioned on proprioception, flow features, tactile features, the contact gate, and the base action chunk:
\begin{equation}
    \mathbf{a}^{\mathrm{r}}_t
    =
    \pi^{\mathrm{r}}_{\theta}
    (
    \mathbf{q}_t,\,
    \mathbf{z}^{f}_t,\,
    \mathbf{g}_t\cdot\mathbf{z}^{\tau}_t,\,
    \mathbf{a}_t,
    \mathbf{a}^{\mathrm{b}}_{t:t+K}
    ),
\end{equation}
where
\begin{equation}
    g_t = \mathbf{1}[m_t > \epsilon_{\mathrm{contact}}]
\end{equation}
is a contact-aware tactile gate and $m_t$ is the marker displacement magnitude. 
The gate prevents the policy from overusing tactile features before contact and encourages tactile-driven corrections only when contact is present. For the other variables, $\mathbf{q}_t$ is the robot proprioception, $\mathbf{z}^f_t$ is the flow representations, $\mathbf{a}_t$ is the current base action, and $\mathbf{a}^{\mathrm{b}}_{t:t+K}$ is the current action chunk, which can provides the temporal reference for residual policy to generate the correction.

For stable real-world learning, we constrain the residual action magnitude:
\begin{equation}
    \mathbf{a}^{\mathrm{r}}_t
    =
    s_t
    \tilde{\mathbf{a}}^{\mathrm{r}}_t,
\end{equation}
where $s_t$ is a scheduled residual scale, defined as $0$--$0.15$. 

The final SAC target is:
\begin{equation}
y_t^{\mathrm{rl}}
=
r_t
+
\gamma
\left[
\min_{i=1,2}
\bar{Q}_{\bar{\eta}_i}
(
\mathbf{q}_{t+1},
\mathbf{z}^{f}_{t+1},
\mathbf{z}^{\tau}_{t+1},
\mathbf{a}^{\mathrm{b}}_{t+1}
+
s_{t+1}\tilde{\mathbf{a}}^{\mathrm{r}}_{t+1}
)
-
\alpha
\log
\pi^{\mathrm{r}}_\theta
(
\tilde{\mathbf{a}}^{\mathrm{r}}_{t+1}
\mid
\mathbf{s}_{t+1}
)
\right].
\end{equation}

In our experiments, $s_t$ is gradually increased during online learning so that early residual updates remain conservative and later updates can make stronger contact corrections. 
This follows the same principle as residual RL methods that keep the learned correction small relative to the base policy to avoid unsafe exploration.

\subsection{Reward Shaping Details}
\label{app:reward}
We use a normalized multi-sensory reward:
\begin{equation}
    r_t
    =
    \frac{
    w_r r^{\mathrm{reach}}_t
    +
    w_g r^{\mathrm{grasp}}_t
    +
    w_f r^{\mathrm{flow}}_t
    -
    w_s r^{\mathrm{safety}}_t
    }
    {w_r+w_g+w_f},
\end{equation}
where $r^{\mathrm{reach}}_t$, $r^{\mathrm{grasp}}_t$, and $r^{\mathrm{flow}}_t$ are normalized to $[0,1]$. 
We use $w_r=1.0$, $w_g=0.5$, $w_f=2.0$, and $w_s=1.0$ for all tasks. 
Because the positive reward is divided by $w_r+w_g+w_f$, the maximum reward is $1$ when all positive components are fully satisfied and the safety penalty is not triggered.

Before grasping, the reaching reward encourages the end-effector to move toward the initial 3D object centroid:
\begin{equation}
    r^{\mathrm{reach}}_t
    =
    1-
    \mathrm{clip}
    \left(
    \frac{
    \|\mathbf{p}^{ee}_t-\mathbf{c}^{3D}_0\|_2
    }
    {d^{\mathrm{reach}}_{\max}},
    0,1
    \right),
\end{equation}
where $d^{\mathrm{reach}}_{\max}=0.15$ m.

After grasping, the flow reward tracks sparse subgoals from the generated object flow:
\begin{equation}
    d^{\mathrm{flow}}_t
    =
    \left\|
    \mathbf{T}^{\mathrm{rel}}_{t_0\rightarrow t}
    -
    \hat{\mathbf{T}}^{\mathrm{rel}}_{t_0\rightarrow \ell}
    \right\|_2,
\end{equation}
\begin{equation}
    r^{\mathrm{flow}}_t
    =
    \frac{\ell_t}{L}
    +
    \frac{1}{L}
    \left(
    1-
    \mathrm{clip}
    \left(
    \frac{d^{\mathrm{flow}}_t}{d^{\mathrm{flow}}_{\max}},
    0,1
    \right)
    \right),
\end{equation}
where $L=30$ is the number of sparse flow subgoals, $\ell_t\in\{0,\dots,L-1\}$ is the number of reached subgoals before the current one, and $d^{\mathrm{flow}}_{\max}=0.05$. 
A flow subgoal is considered reached when $d^{\mathrm{flow}}_t<\epsilon_{\mathrm{flow}}$, where $\epsilon_{\mathrm{flow}}=0.03$.

The tactile grasp reward encourages stable contact:
\begin{equation}
    r^{\mathrm{grasp}}_t
    =
    \mathbf{1}
    \left[
    \frac{1}{|\Omega|}
    \sum_{u\in\Omega}
    D^{\tau}_t(u)
    >
    \epsilon_{\mathrm{depth}}
    \right],
\end{equation}
where $D^{\tau}_t$ is the tactile depth image, $\Omega$ is the valid tactile region, and $\epsilon_{\mathrm{depth}}=0.10$. 
For marker-based tactile sensing, we instead compute the average marker displacement magnitude:
\begin{equation}
    m_t
    =
    \frac{1}{|\mathcal{M}|}
    \sum_{j\in\mathcal{M}}
    \left\|
    \Delta \mathbf{u}^{j}_t
    \right\|_2,
\end{equation}
where $\mathcal{M}$ is the set of detected markers and $\Delta \mathbf{u}^{j}_t$ is the displacement of marker $j$ from its reference position. 
Marker-based contact is detected when $m_t>\epsilon_{\mathrm{contact}}$, where $\epsilon_{\mathrm{contact}}=1.5$ pixels.

The safety penalty detects excessive contact:
\begin{equation}
    r^{\mathrm{safety}}_t
    =
    \mathbf{1}
    [
    m_t>\epsilon_{\mathrm{safety}}
    ],
\end{equation}
where $\epsilon_{\mathrm{safety}}=8.0$ pixels. 
If the safety penalty is triggered, we terminate the episode and reset the environment.

\subsection{Training and Inference Details}
We train all supervised models on a single NVIDIA RTX A6000 GPU. This includes flow-generator fine-tuning, the flow-policy training, and the supervised training or fine-tuning of ACT, DP, and $\pi_{0.5}$. 

For real-world RL, we follow a similar asynchronous training setup as prior real-world RL systems such as HIL-SERL~\cite{serl} and WSRL~\cite{effrl}: policy execution and network optimization run in parallel, so the robot can continue collecting interaction data while the learner updates the critic, residual policy, and tactile encoder from the replay buffer. 
During both warm-start and online RL, we update the trainable networks every $500$ optimization steps. 

In the warm-start stage, the residual action is set to zero, where only the critic and tactile encoder are optimized. After each warm-start episode, we run ControlTac~\cite{controltac} augmentation on the collected contact segments and add the augmented tactile transitions to the replay buffer. ControlTac augmentation is executed asynchronously together with flow-generator inference, so tactile augmentation does not block robot execution. 

For deployment-time inference, the robot runs the frozen visual base policy and the residual tactile policy on a single NVIDIA RTX 5080 GPU.

\subsection{Hyperparameters}
\label{app:hyperparameters}

Unless otherwise specified, we use the same hyperparameters across all tasks and base policies. 
All actions are represented in the robot end-effector frame and normalized to $[-1,1]$ before being passed to the policy. 
The control frequency is $10$ Hz, and each episode has a maximum length of $300$ control steps.

\begin{table}[h]
    \centering
    \caption{Implementation hyperparameters used in \ourframework.}
    \label{tab:app_hyperparams}
    \resizebox{0.92\linewidth}{!}{
    \begin{tabular}{llc}
        \toprule
        \textbf{Component} & \textbf{Hyperparameter} & \textbf{Value} \\
        \midrule
        \multicolumn{3}{l}{\textit{Data collection and retargeting}} \\
        Camera & Third-view RGB frame rate & $30$ Hz \\
        Quest & Quest hand tracking rate & $60$ Hz \\
        Pose smoothing & Translation smoothing window & $5$ frames \\
        Finger retargeting & Closed/open hand distances $(d_{\min},d_{\max})$ & $(5,12)$ cm \\
        Finger retargeting & Gripper range $[a_{\min},a_{\max}]$ & $[0,1]$ \\
        Finger retargeting & Aperture scale/offset $(\alpha,\beta)$ & $(1.0,0.0)$ \\
        Grasp detection & Aperture / velocity threshold & $0.35$ / $-0.02$ m/s \\
        \midrule
        \multicolumn{3}{l}{\textit{Flow generator and flow policy}} \\
        Object keypoints & Number of tracked keypoints $N$ & $32$ \\
        Flow generator & Demonstrations per task & $50$ \\
        Flow generator & LoRA rank / alpha & $16 / 32$ \\
        Flow generator & Learning rate / batch size / epochs & $1\times10^{-4}$ / $8$ / $500$ \\
        Flow loss & Smoothness weight $\lambda_{\mathrm{smooth}}$ & $0.1$ \\
        Sparse flow & Number of sparse subgoals $L$ & $30$ \\
        Flow policy & Action chunk horizon $K$ & $8$ \\
        \midrule
        \multicolumn{3}{l}{\textit{Warm-start and tactile representation}} \\
        Warm-start & Duration & $12$ min \\
        Replay buffer & Maximum size & $1\times10^{4}$ transitions \\
        Tactile latent & Latent dimension $d_{\tau}$ & $32$ \\
        Encoder loss & Reconstruction weight $\lambda_{\mathrm{rec}}$ & $0.1$ \\
        Encoder update & Contact-only update & Yes \\
        \midrule
        \multicolumn{3}{l}{\textit{ControlTac trajectory-level augmentation}} \\
        ControlTac & Augmented trajectories per real trajectory & $2$ \\
        ControlTac & Force perturbation $\Delta f$ & $\mathcal{U}[-3,10]$ \\
        ControlTac & Pose/control offset candidates & $\{-1,-0.5,0,0.5,1\}$ \\
        ControlTac & Generation granularity & trajectory-level \\
        \midrule
        \multicolumn{3}{l}{\textit{Online residual RL}} \\
        RL algorithm & Backbone & SAC \\
        SAC & Discount / target update $(\gamma,\tau)$ & $(0.99,0.005)$ \\
        SAC & Actor / critic learning rate & $3\times10^{-4}$ \\
        SAC & Encoder learning rate & $1\times10^{-4}$ \\
        SAC & Batch size / updates per step & $256$ / $1$ \\
        Policy
        Residual scale & Schedule $s_t$ & $0.05 \rightarrow 0.10 \rightarrow 0.15$ \\
        Residual scale & Schedule interval & every $500$ online steps \\
        \bottomrule
    \end{tabular}
    }
\end{table}\clearpage

\begin{table}[t]
    \centering
    \caption{Reward, contact-gate, and safety hyperparameters. The positive reward is normalized to $[0,1]$, while the safety term is applied as an additional penalty.}
    \label{tab:app_reward_hyperparams}
    \resizebox{0.85\linewidth}{!}{
    \begin{tabular}{llc}
        \toprule
        \textbf{Component} & \textbf{Hyperparameter} & \textbf{Value} \\
        \midrule
        \multicolumn{3}{l}{\textit{Reward weights and normalization}} \\
        Reward & Reaching weight $w_r$ & $1.0$ \\
        Reward & Grasp/contact weight $w_g$ & $0.5$ \\
        Reward & Flow weight $w_f$ & $2.0$ \\
        Reward & Safety penalty weight $w_s$ & $1.0$ \\
        Reward & Positive reward normalization & $w_r+w_g+w_f=3.5$ \\
        Reward & Positive reward range & $[0,1]$ \\
        Safety reward & Safety penalty range & $[-1,0]$ \\
        \midrule
        \multicolumn{3}{l}{\textit{Reaching and flow reward}} \\
        Reaching reward & Distance normalization $d^{\mathrm{reach}}_{\max}$ & $0.15$ m \\
        Flow reward & Number of sparse subgoals $L$ & $30$ \\
        Flow reward & Flow error normalization $d^{\mathrm{flow}}_{\max}$ & $0.05$ \\
        Flow reward & Flow subgoal threshold $\epsilon_{\mathrm{flow}}$ & $0.03$ \\
        Subgoal switching & Policy subgoal threshold $\epsilon_{\mathrm{subgoal}}$ & $0.03$ \\
        \midrule
        \multicolumn{3}{l}{\textit{Contact detection, tactile gate, and safety}} \\
        Tactile gate & Marker gate threshold $\epsilon_{\mathrm{gate}}$ & $1.5$ px \\
        Contact detection & Marker contact threshold $\epsilon_{\mathrm{contact}}$ & $1.5$ px \\
        Contact detection & Tactile depth threshold $\epsilon_{\mathrm{depth}}$ & $0.10$ \\
        Safety reset & Marker safety threshold $\epsilon_{\mathrm{safety}}$ & $8.0$ px \\
        Safety reset & Termination condition & $m_t>\epsilon_{\mathrm{safety}}$ \\
        \bottomrule
    \end{tabular}
    }
\end{table}

\section{Experimental Settings}
\label{app:setting}

\begin{figure*}[!h]
    \centering
    \begin{minipage}[c]{0.38\textwidth}
\paragraph{Peg-in-Hole Task.}
In this task, the robot grasps a cylindrical peg and inserts it into a target receptacle placed at different tabletop locations. The task evaluates whether the policy can generalize the reaching and grasping motion across spatial variations while still performing precise contact-rich insertion. More details have been shown in Fig.~\ref{fig:peg}, where the tolerance of the insertion task is 5mm. Although the visual base policy can usually bring the peg close to the hole, small pose errors during insertion often lead to jamming or failed contact. Tactile feedback is therefore critical for detecting contact, correcting local misalignment, and maintaining a stable grasp during the final insertion stage.
    \end{minipage}
    \hfill
    \begin{minipage}[c]{0.6\textwidth}
        \centering
        \includegraphics[width=\linewidth]{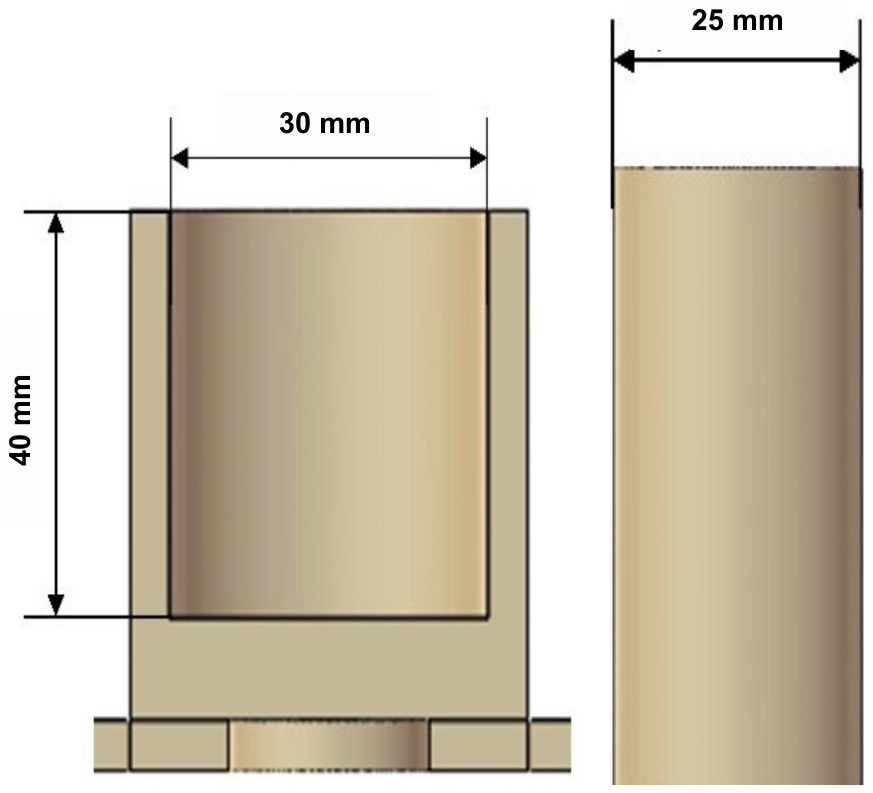}
        \caption{Setting of Peg-in-Hole task.}
        \label{fig:peg}
    \end{minipage}
    \vspace{-10pt}
\end{figure*}

\begin{figure*}
    \centering
    \includegraphics[width=0.7\textwidth]{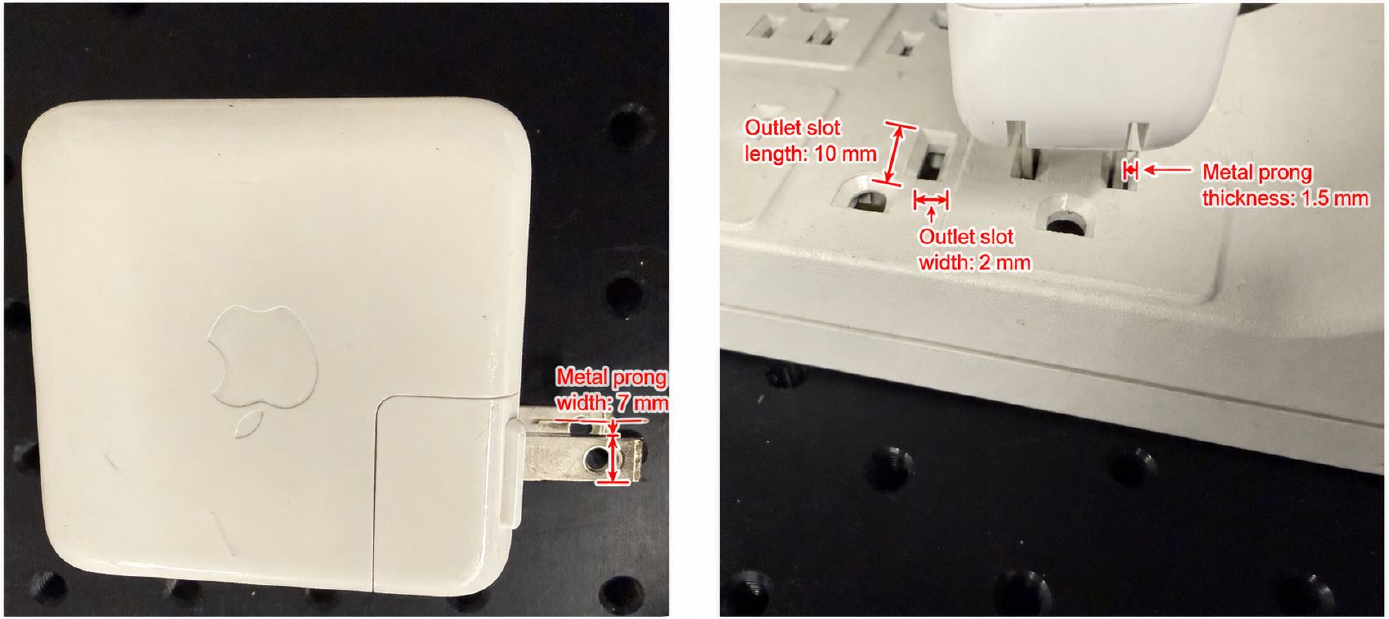}
        \caption{Setting of Charger Insertion task.}
        \label{fig:charger}
    \vspace{-10pt}
\end{figure*}
\paragraph{Charger Insertion Task.}
The robot has been grasped a charger and plan to insert it into a phone charging port. This task is substantially more precise than standard insertion because both the charger head and the socket tolerance are very small. As shown in Fig.~\ref{fig:charger}, the metal prong width is only about $7$ mm, while the corresponding outlet slot length is about $10$ mm. For the smaller side, the metal prong thickness is about $1.5$ mm and the socket width is about $2$ mm, leaving only a narrow clearance for successful insertion. As a result, visual guidance alone often brings the charger near the port but cannot reliably resolve the final contact alignment. The robot must use tactile feedback to explore contact, detect lateral misalignment, and make small corrective motions before insertion. To improve robustness, we apply small domain randomization for the initial robot position during both training and inference by perturbing the initial pose within $5$ cm along the $x$, $y$, and $z$ directions and within $10^\circ$ around the vertical axis. 
Since the charger is already grasped and the task is challenging, we restrict the action space of the residual policy to translational corrections in $x$, $y$, and $z$ and rotational correction in $yaw$.

\begin{figure*}[!h]
    \centering
    \begin{minipage}[c]{0.5\textwidth}
\paragraph{Cap Opening Task.}
In this task, the robot grasps a cap opener and uses it to open a bottle cap. 
Unlike pure insertion tasks, this setting requires maintaining a precise tool pose while the contact geometry changes dynamically during levering. The opener must first establish contact with the cap edge, then preserve this contact while applying a rotational motion to lift the cap. Failure modes include slipping off the cap, contacting the wrong region, or applying force with an incorrect tool orientation. 
This task therefore tests whether tactile residual adaptation can stabilize dynamic contact and correct the tool pose during forceful interaction. Similar to Charger Insertion, 
    \end{minipage}
    \hfill
    \begin{minipage}[c]{0.48\textwidth}
        \centering
        \includegraphics[width=\linewidth]{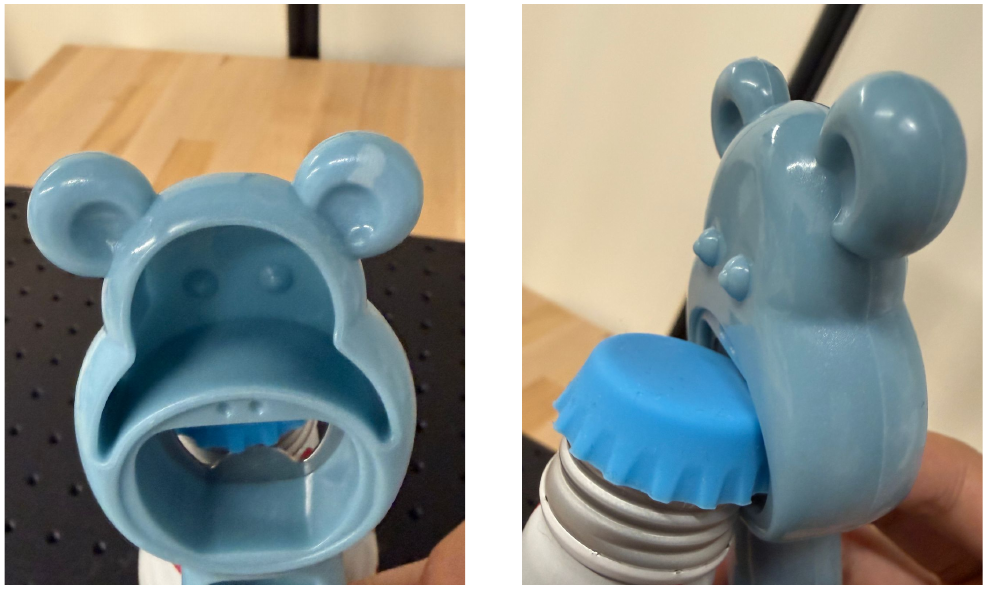}
        \caption{Setting of Cap Opening task.}
        \label{fig:cap}
    \end{minipage}
    \vspace{-10pt}
\end{figure*}
we apply small domain randomization for the initial robot position during both training and inference by perturbing the initial pose within $5$ cm along the $x$, $y$, and $z$ directions and within $10^\circ$ around the vertical axis. The setting is visualized in Fig.~\ref{fig:cap}.

\begin{figure}[h]
    \centering
    \includegraphics[width=0.7\textwidth]{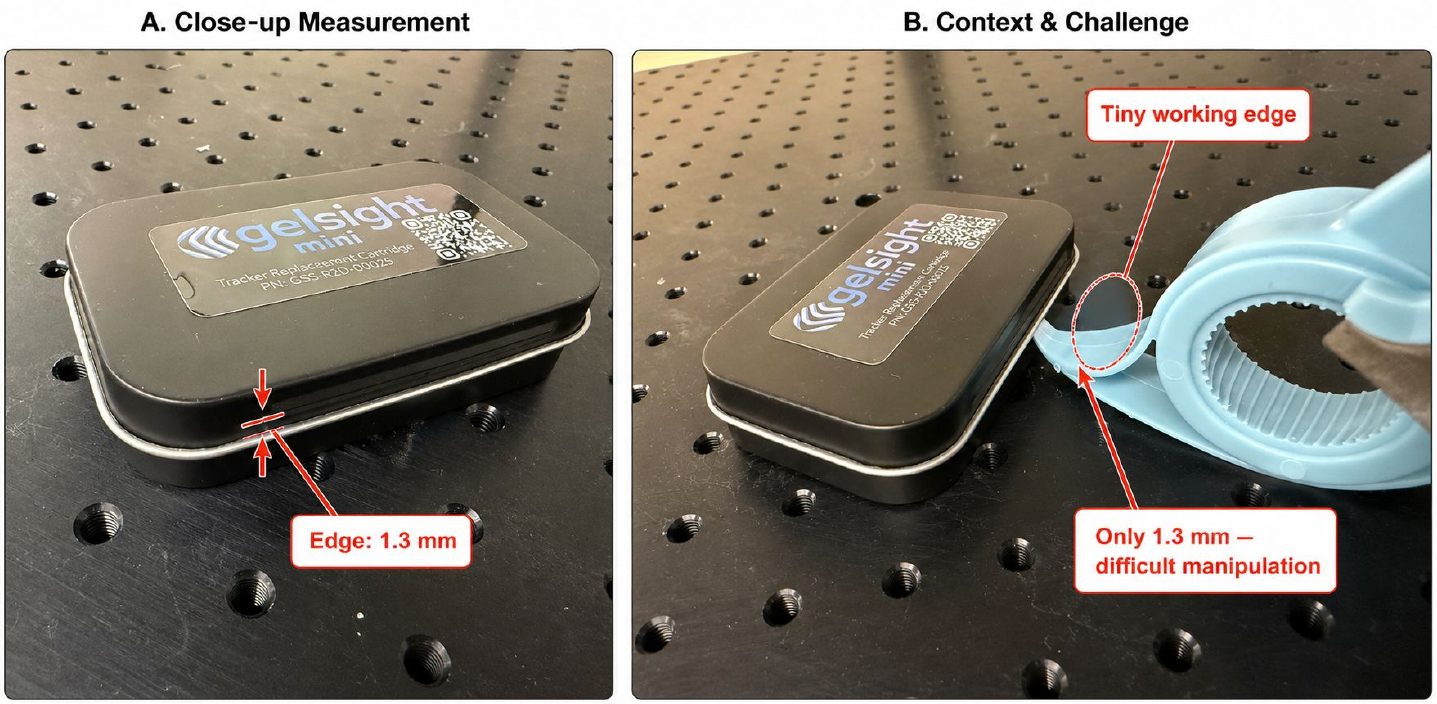}
        \caption{Setting of Box Opening task.}
        \label{fig:box}
    \vspace{-10pt}
\end{figure}

\paragraph{Box Opening Task.}
The robot uses a lever-type opener to open the GelSight box by catching and lifting a very thin edge. 
This task is particularly challenging because the available working edge is only about $1.3$ mm thick, as shown in Fig.~\ref{fig:box}. 
The robot must precisely align the opener with this narrow edge, establish contact without slipping, and maintain the contact while executing a dynamic levering motion. 
Small visual or pose errors can easily cause the opener to miss the edge or slide away from it. 
Thus, the task strongly depends on tactile feedback for detecting edge contact, maintaining stable engagement, and correcting the manipulation trajectory during opening. Same as the other two tasks, we apply small domain randomization for the initial robot position during both training and inference by perturbing the initial pose within $5$ cm along the $x$, $y$, and $z$ directions and within $10^\circ$ around the vertical axis.

\newpage
\section{Additional Experiments}
\label{app:experiment}
\subsection{Analysis of Collected Data and Base Policies}~\label{app:analysis}
In this section, we want to analyze the trajectory quality of the base policies and the data. To evaluate their quality, we propose to use the smoothness of the tracked keypoints centroid on the object, since all sources share this representation. Following prior work~\cite{sail}, we use two complementary smoothness metrics: SPARC and LDLJ.

SPARC measures the arc length of the normalized Fourier magnitude spectrum of the speed profile, capturing high-frequency oscillations in the trajectory:
\begin{equation}
    \mathrm{SPARC} \triangleq 
    - \int_{0}^{\omega_c}
    \left[
    \left(\frac{1}{\omega_c}\right)^2 +
    \left(\frac{d\hat{V}(\omega)}{d\omega}\right)^2
    \right]^{1/2} d\omega ,
\end{equation}
where $\hat{V}(\omega)=V(\omega)/V(0)$ is the normalized magnitude spectrum of $v_t$, and $\omega_c$ is an adaptive cutoff frequency determined by a spectrum amplitude threshold. 
A larger SPARC value, i.e., closer to zero, indicates a smoother trajectory with fewer high-frequency components.

We further use LDLJ, which measures smoothness in the time domain by penalizing abrupt changes in acceleration:
\begin{equation}
    \mathrm{LDLJ} \triangleq
    -\log\left(
    \frac{(t_2-t_1)^5}{v_{\mathrm{peak}}^2}
    \int_{t_1}^{t_2}
    \left\|
    \frac{d^2 v(t)}{dt^2}
    \right\|_2^2 dt
    \right),
\end{equation}
where $t_1$ and $t_2$ denote the start and end time of the trajectory, and $v_{\mathrm{peak}}=\max_{t\in[t_1,t_2]} v(t)$ is the peak speed. 
LDLJ is dimensionless and penalizes jerky motions caused by rapid acceleration and deceleration changes. 
We report mean SPARC and LDLJ over successful rollouts, denoted as M-SPARC and M-LDLJ, where larger values indicate smoother motion.

We compare four trajectory sources that share the same object-centric keypoint representation: human demonstrations, teleoperation demonstrations, the policy trained from human demonstrations, and the policy trained from teleoperation demonstrations. 
In addition to quantitative metrics, we visualize successful trajectories from the same initial setting for all four groups, allowing direct comparison of the trajectory shape and contact-stage behavior.

\begin{figure*}[h]
    \centering
    \begin{minipage}[c]{0.48\textwidth}
        \centering
        \includegraphics[width=\linewidth]{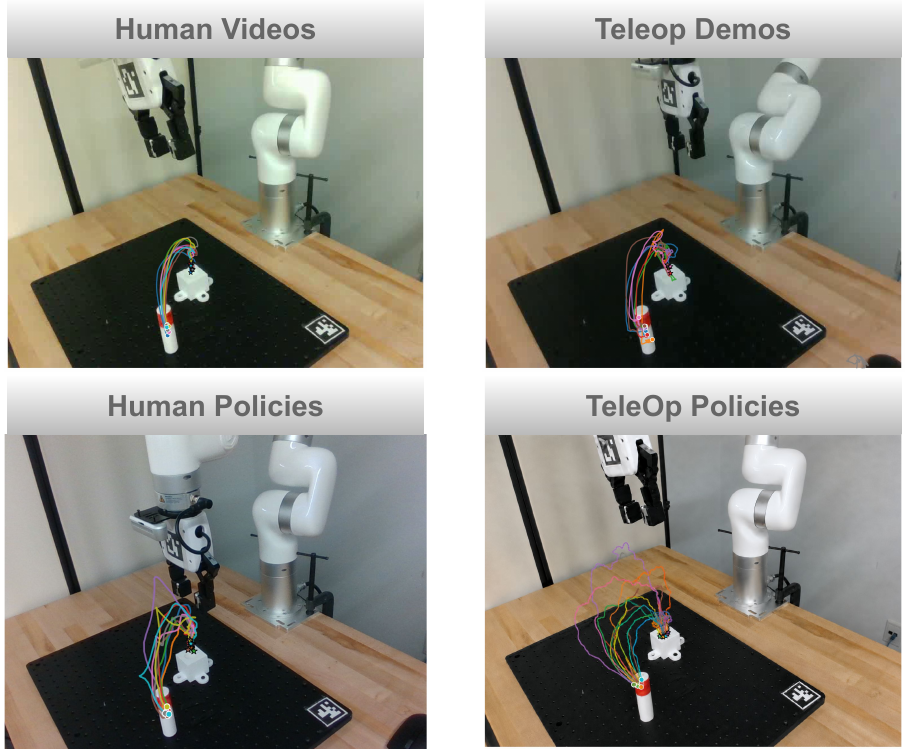}
        \caption{Visualization of centroid trajectories.}
        \label{fig:centroids}
    \end{minipage}
    \hfill
    \begin{minipage}[c]{0.5\textwidth}
        \centering
        \captionof{table}{Trajectory smoothness comparison using the centroid of tracked object keypoints.}
        \label{tab:smoothness}
        \resizebox{\linewidth}{!}{
        \begin{tabular}{lcc}
            \toprule
            \textbf{Group} & \textbf{M-SPARC} $\uparrow$ & \textbf{M-LDLJ} $\uparrow$ \\
            \midrule
            Human Demos    & $-2.889$ & $-18.03$ \\
            Teleop Demos   & $-7.320$ & $-19.70$ \\
            Human Policy   & $-2.664$ & $-17.29$ \\
            TeleOp Policy  & $-9.521$ & $-20.89$ \\
            \bottomrule
        \end{tabular}
        }
    \end{minipage}
    \vspace{-1em}
\end{figure*}

Table~\ref{tab:smoothness} and Fig.~\ref{fig:centroids} shows that human demonstrations and the human policy produce substantially smoother object trajectories than their teleoperation counterparts. Human demonstrations achieve an M-SPARC of $-2.889$ and M-LDLJ of $-18.03$, while teleoperation demonstrations drop to $-7.320$ and $-19.70$, indicating more oscillations and jerkier motion. This gap becomes larger after policy learning: the human policy remains smooth with M-SPARC $-2.664$ and M-LDLJ $-17.29$, whereas the teleoperation policy further degrades to M-SPARC $-9.521$ and M-LDLJ $-20.89$.

These results suggest that teleoperation data is less suitable for contact-rich manipulation. 
During teleoperation, the human operator cannot directly feel contact forces at the object, making precise contact exploration and insertion difficult. 
As a result, teleoperation trajectories often contain noisy corrections and jerky motions, especially during the contact-rich phase. 
When a policy imitates such data, these artifacts can be amplified, producing unstable contact behavior and making tactile residual refinement more difficult. 
This also indicates that collecting teleoperation data for more challenging contact-rich tasks can become impractical, as the demonstrations themselves may not provide sufficiently smooth and reliable motion priors.

\subsection{Ablation Studies}
\vspace{-0.15cm}
\begin{figure}[h]
    \centering
    \begin{minipage}[c]{0.62\textwidth}
        \paragraph{Reward Shaping }We conduct ablation study for our multi-sensory reward shaping method in the Peg-in-Hole task. As shown in Fig.~\ref{fig:reward}, removing any component of our reward design leads to slower learning and lower final performance, demonstrating the importance of multi-sensory reward shaping. The reaching reward provides guidance for grasping the object, while the dense flow reward further offers object-centric motion supervision during insertion. In addition, the tactile reward captures grasping signals and adding penalties for unexpected behaviors.
    \end{minipage}
    \hfill
    \begin{minipage}[c]{0.35\textwidth}
        \centering
    \includegraphics[width=\textwidth]{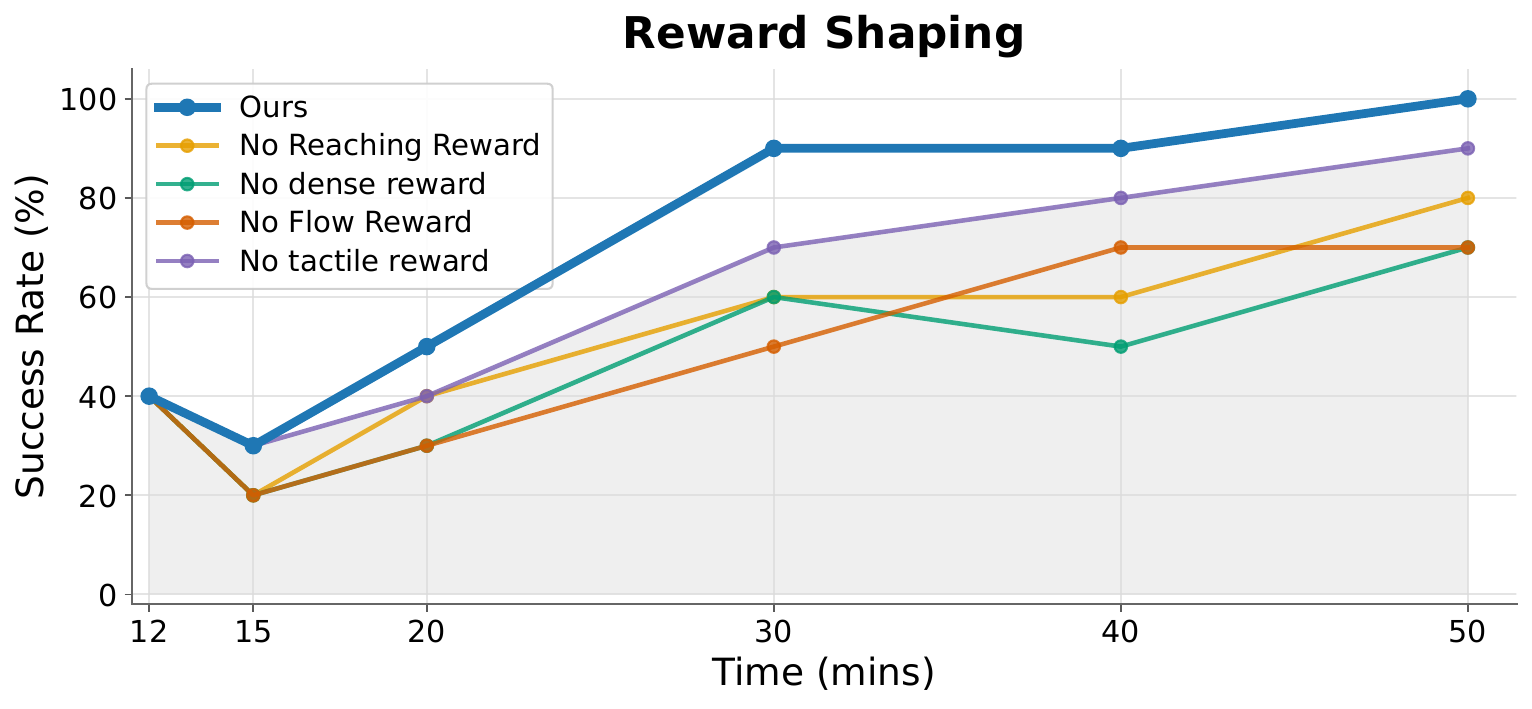}
    \caption{Ablation Study of the Multi-sensory Reward Shaping}
    \label{fig:reward}
    \end{minipage}
\end{figure}
\vspace{-0.15cm}
\begin{figure}[h]
    \centering
    \begin{minipage}[c]{0.62\textwidth}
        \paragraph{Residual  Policy Design }We further ablate our residual policy design in Fig.~\ref{fig:policy}. Our method uses trajectory-level keypoint guidance, which is temporally sparser than per-step keypoint conditions but still providing a structured motion prior. As shown in the figure, per-step keypoint will over-constrain the policy, which leads the base policy such as ACT failed because of its highly varied actions. In contrast, trajectory-level keypoints leaves enough reactivity for residual RL to refine local contact-rich behaviors. We also find that keypoint-conditioned policy is more effective than using visuo-tactile observations, since keypoints provide a compact object-centric guidance. Finally, contact-aware gating also improves performance by reducing unnecessary corrections in non-contact stage and enabling stronger tactile-driven corrections during contact.
    \end{minipage}
    \hfill
    \begin{minipage}[c]{0.35\textwidth}
        \centering
    \includegraphics[width=\textwidth]{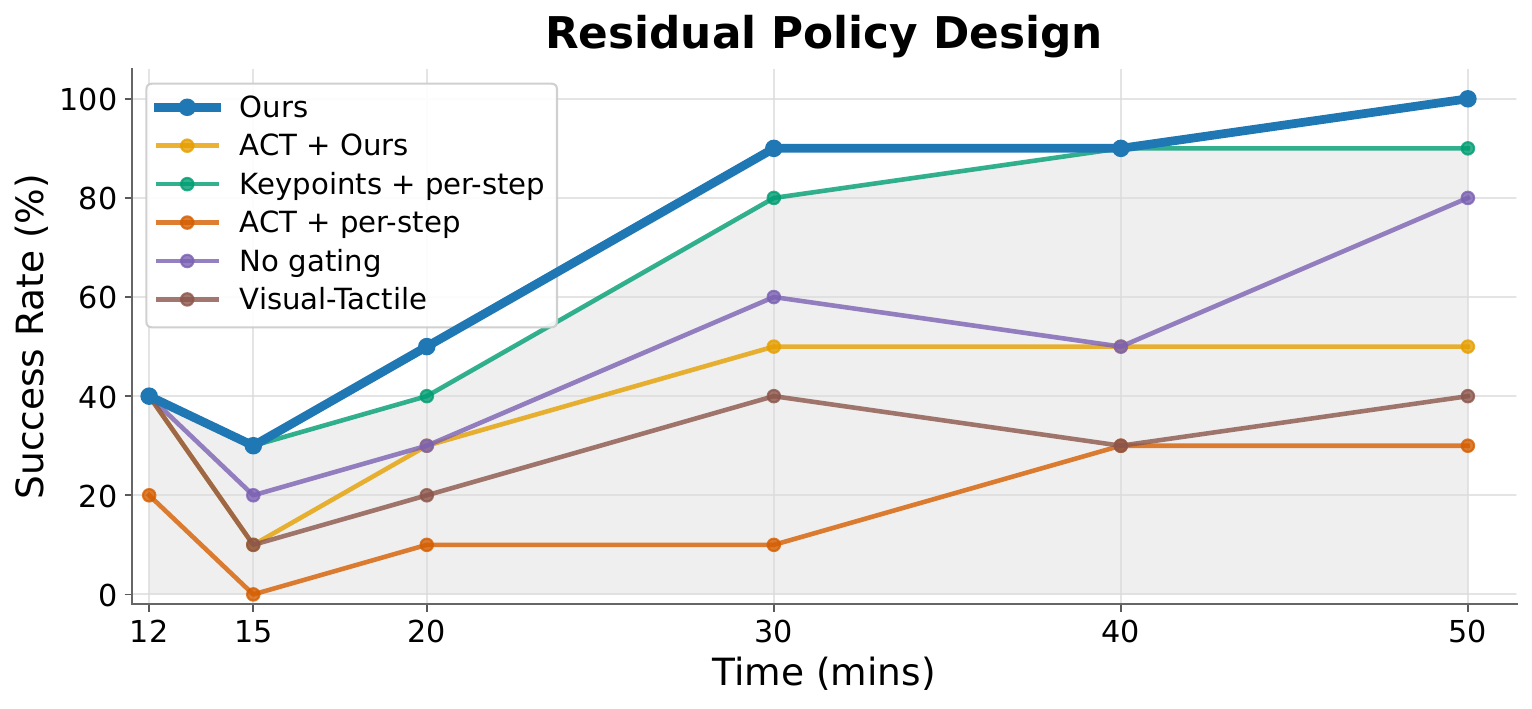}
    \caption{Ablation between our residual policy design and per-step keypoint reaching policy, visuo-tactile policy, or policy after removing contact-aware gating.}
    \label{fig:policy}
    \end{minipage}
\end{figure}
\begin{figure}[h]
    \centering
    \begin{minipage}[c]{0.62\textwidth}
        \paragraph{Warm-Start Strategies }We then conduct ablation study for the design choices in our warm-start stage. As shown in the  Fig.~\ref{fig:warm-start}, removing warm-start optimization on the tactile encoder and critic leads to unstable early stage and weak overall performance, which indicates that optimizing the tactile representation and critic during warm-start is important for stable residual RL. We also observe that removing ControlTac~\cite{controltac} also results in unstable performance, highlighting that tactile augmentation in the warm-start stage can refine the tactile encoder more efficiently. In addition, removing the critic loss or the reconstruction loss both degrades final performance, showing that the two objectives play complementary roles in extracting task-level features and restoring tactile structures.
    \end{minipage}
    \hfill
    \begin{minipage}[c]{0.35\textwidth}
        \centering
    \includegraphics[width=\textwidth]{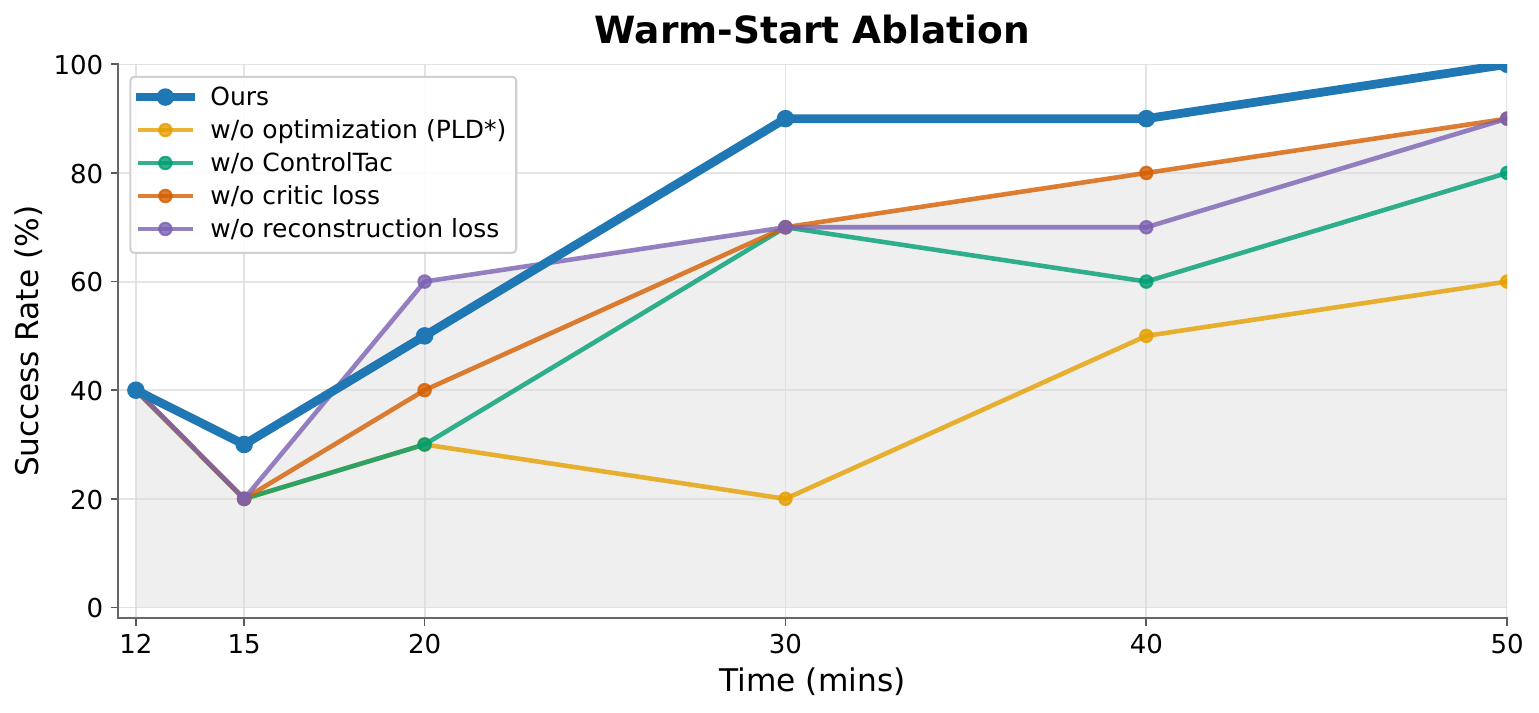}
    \caption{Ablation study of the warm-start stage in showing the importance of optimization of both tactile encoder and critic.}
    \label{fig:warm-start}
    \end{minipage}
\end{figure}

\begin{figure}[h]
    \centering
    \begin{minipage}[c]{0.62\textwidth}
        \paragraph{Action Space Design }We further ablate our action space design in Fig.~\ref{fig:action}. As shown in the figure, removing the scheduler leads to slower learning and lower final performance, suggesting that stage-aware action scheduling is important for balancing exploration and stable refinement during contact-rich manipulation. We also observe that removing action scaling significantly hurts performance, indicating that unconstrained residual corrections are harder to optimize and can easily destabilize training. In addition, using a larger action scale of $0.5$ with scheduler improves over removing action scaling, but still underperforms our full design, showing that $0,15$ is an appropriate ratio. Overall, these results validate that both the scheduler and action scaling are important for making residual actions more stable and sample-efficient.
    \end{minipage}
    \hfill
    \begin{minipage}[c]{0.35\textwidth}
        \centering
    \includegraphics[width=\textwidth]{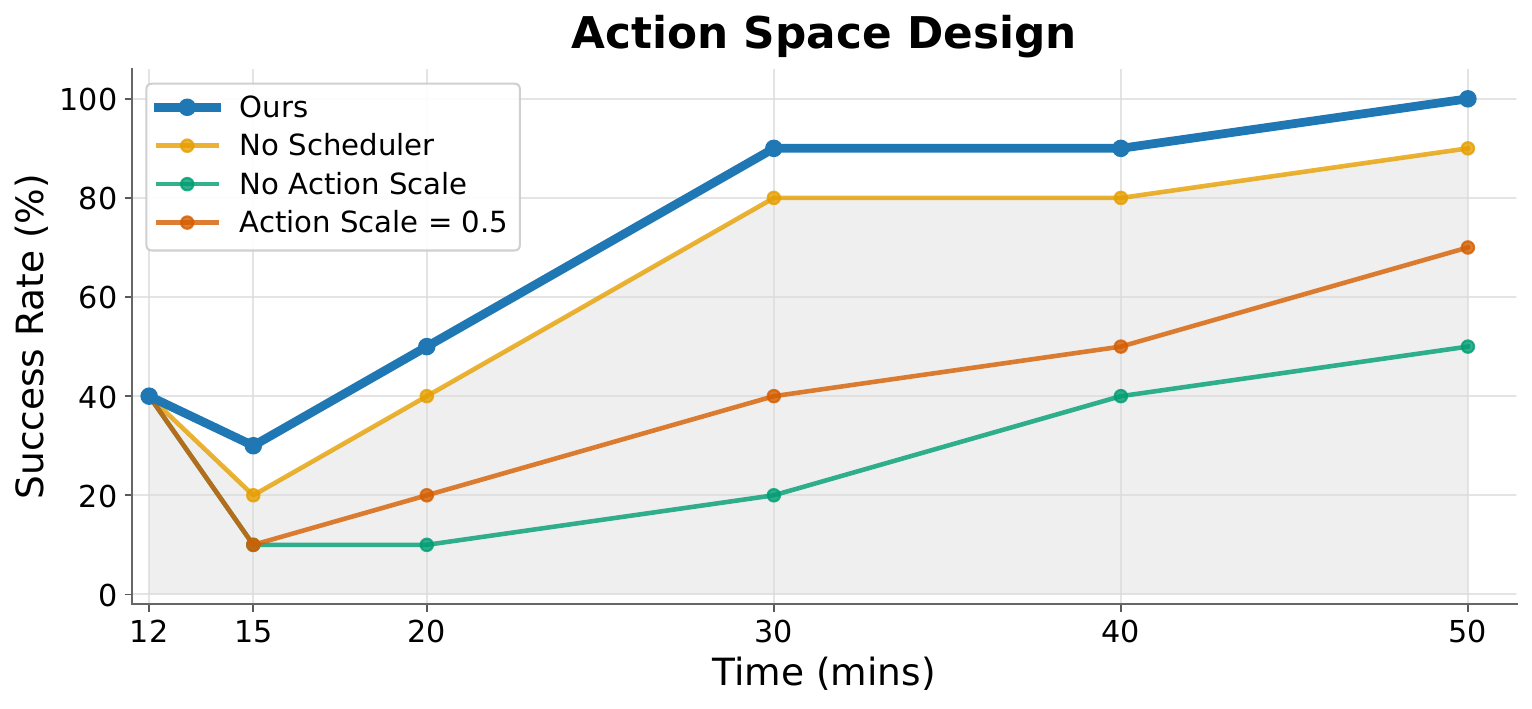}
    \caption{Ablation study of the action space in showing the importance of scheduler and the scale number.}
    \label{fig:action}
    \end{minipage}
\end{figure}

\newpage
\section{Failure Cases}
\label{app:failure}
\begin{figure*}[!h]
    \centering
    \includegraphics[width=0.9\textwidth]{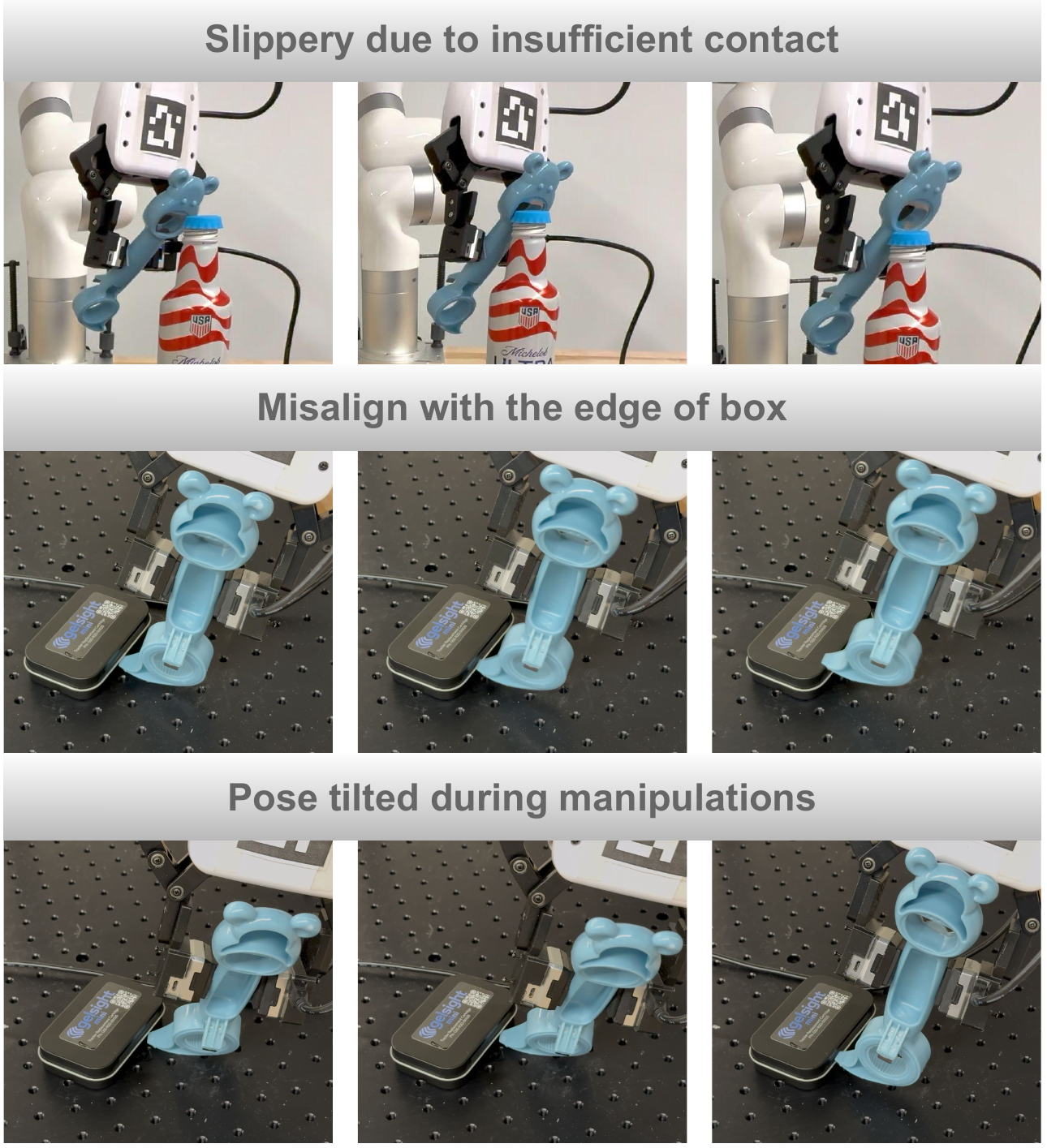}
    \caption{Visualization of the failure cases on \textbf{Cap Opening} and \textbf{Box Opening}.}
    \label{fig:failure}
    \vspace{-10pt}
\end{figure*}

In this section, we further analyze representative failure cases in the \textbf{Cap Opening} and \textbf{Box Opening} tasks, as shown in Fig.~\ref{fig:failure}. 
First, even when the object reaches the cap, the contact force may be insufficient or imprecisely applied, leading to slipping rather than successful opening. 
Second, the policy may fail to align the object with the box edge, causing the contact point to miss the effective opening region. 
Third, the object pose can become tilted during manipulation, which changes the contact geometry and prevents the robot from applying force in the desired direction. 
These failures suggest that cap opening and box opening require not only reaching the correct location, but also maintaining a stable object pose and generating accurate contact-rich corrective motions.

\newpage
\section{Additional plots}
\begin{figure*}[!h]
    \centering
    \includegraphics[width=0.9\textwidth]{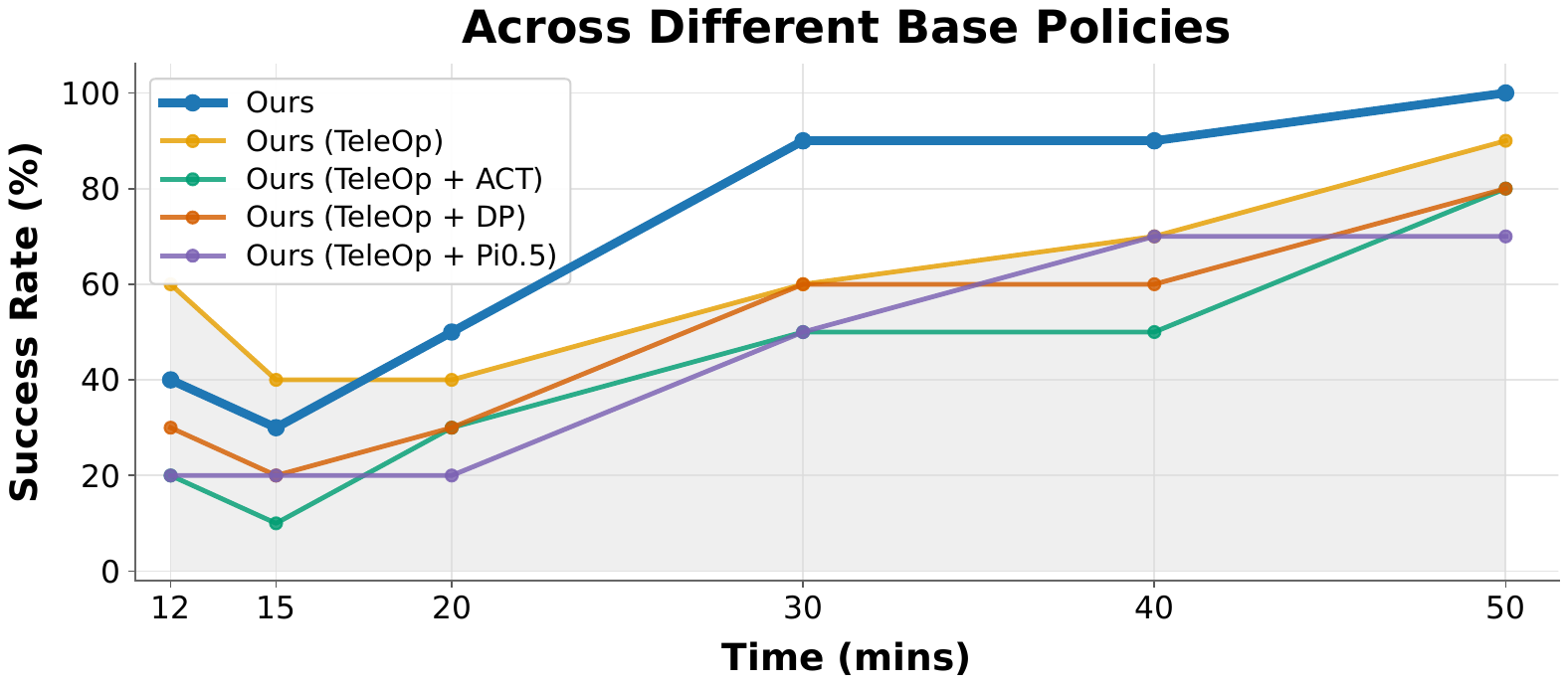}
    \caption{The training success rates for different policies, which is shown in Sec.~\ref{sec:general}.}
    \label{fig:policies_plot}
    \vspace{-10pt}
\end{figure*}

\begin{figure*}[!h]
    \centering
    \includegraphics[width=0.9\textwidth]{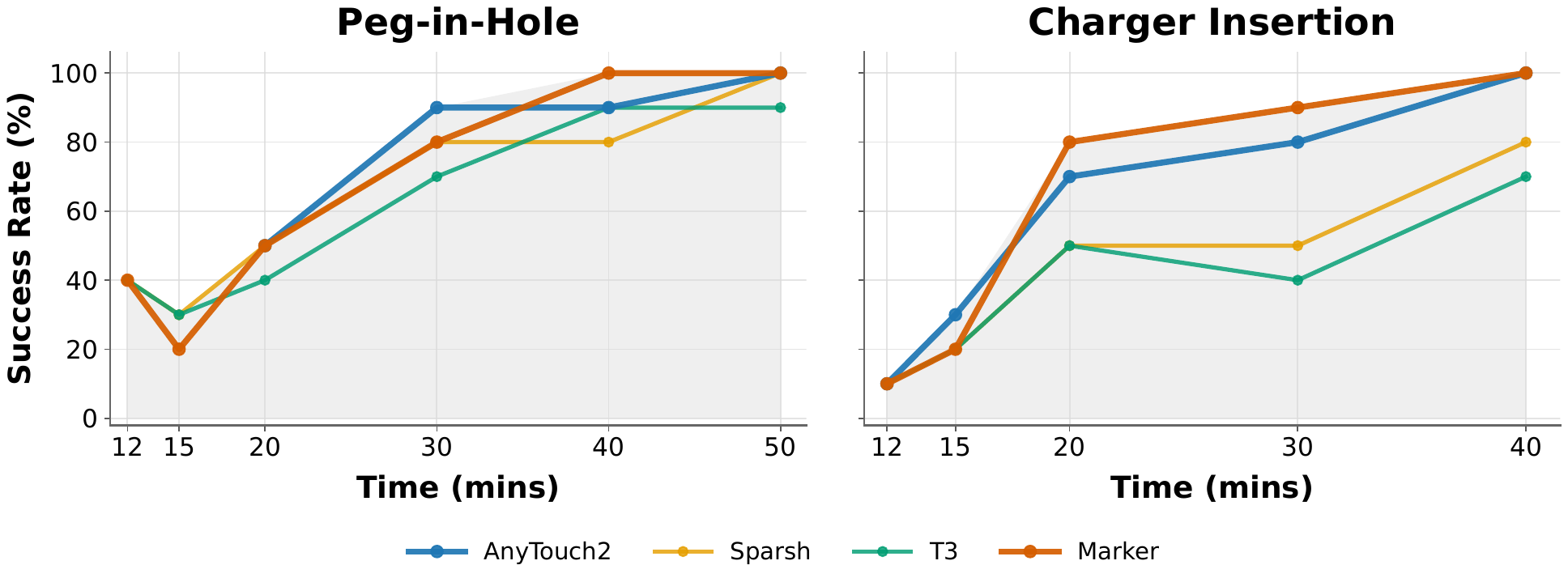}
    \caption{The training success rates for different policies, which is shown in Sec.~\ref{sec:tact}.}
    \label{fig:tact_plot}
    \vspace{-10pt}
\end{figure*}

In this section, we visualize the training curve of the success rates across different visual base policies and tactile representations, which are additional results of Sec.~\ref{sec:general} and Sec.~\ref{sec:tact}. The plots are shown in Fig.~\ref{fig:policies_plot} and Fig.~\ref{fig:tact_plot}.

\end{document}